\documentclass[letterpaper, 10 pt, conference]{ieeeconf}  
\IEEEoverridecommandlockouts                              
\usepackage{graphicx}
\usepackage{tikz}
\usepackage{atbegshi}
\usepackage{amsmath,amssymb}
\usepackage{multirow}
\usepackage{pifont}  
\usepackage{subcaption} 
\usepackage{stfloats}
\usepackage{hyperref}
\usepackage[noadjust]{cite}
\title{\LARGE \bf
LLM-Based Generalizable Hierarchical Task Planning and Execution for Heterogeneous Robot Teams with Event-Driven Replanning.}
\author{Suraj Borate$^{1}$, Bhavish Rai B$^{2}$, Vipul Pardeshi$^{3}$ and Madhu Vadali$^{4}$
\thanks{$^{1}$Suraj Borate is a PhD scholar in the Department of Mechanical Engineering, IIT Gandhinagar, Gujarat, India.
        {\tt\small surajb@iitgn.ac.in}}%
\thanks{$^{2}$Bhavish Rai B is B.Tech graduate from Sahyadri College of Engineering and Management, Adyar, Karnataka, India.
        {\tt\small bhavishraib@gmail.com}}%
\thanks{$^{3}$Vipul Pardeshi is with Vishwakarma Institute of Information Technology, Pune, Maharashtra, India .
        {\tt\small pardeshivipul18@gmail.com}}%
\thanks{$^{4}$Madhu Vadali is an Associate Professor in the Department of Mechanical Engineering, IIT Gandhinagar, Gujarat, India.
        {\tt\small madhu.vadali@iitgn.ac.in}}%
}

\begin{document}
\AtBeginShipout{%
  \AtBeginShipoutUpperLeft{%
    \begin{tikzpicture}[remember picture,overlay]
      \node[anchor=north east] at ([xshift=-5mm,yshift=-5mm]current page.north east)
      {\small\bfseries Preprint};
    \end{tikzpicture}
  }
}

\maketitle

\thispagestyle{empty}
\pagestyle{empty}

\begin{abstract}
This paper introduces CoMuRoS (Collaborative Multi-Robot System), a generalizable hierarchical architecture for heterogeneous robot teams that unifies centralized deliberation with decentralized execution, and supports event-driven replanning. A Task Manager LLM interprets natural-language goals, classifies tasks, and allocates subtasks using static rules plus dynamic contexts (task, history, robot/task status, and events). Each robot runs a local LLM that composes executable Python code from primitive skills (ROS2 nodes, policies), while onboard perception (VLMs/image processing) continuously monitors events and classifies them into relevant or irrelevant to the task. Task failures or user intent changes trigger replanning, allowing robots to assist teammates, resume tasks, or request human help.

Hardware studies demonstrate autonomous recovery from disruptive events, filtering of irrelevant distractions, and tightly coordinated transport with emergent human–robot cooperation (e.g., multirobot collaborative object recovery: success rate 9/10; coordinated transport: 8/8; human-assisted recovery: 5/5). Simulation studies show intention-aware replanning. A curated textual benchmark spanning 22 scenarios (3 tasks each; around 20 robots) evaluates task allocation, classification, IoU, executability, and correctness, with high average scores (e.g., correctness up to 0.91) across multiple LLMs; a separate replanning set (5 scenarios) achieves 1.0 correctness. Compared with prior LLM-based systems, CoMuRoS uniquely demonstrates runtime, event-driven replanning on physical robots, delivering robust, flexible multi-robot and human–robot collaboration.
\end{abstract}
\vspace{-2mm}
\section{INTRODUCTION}
Multi-robot systems can accomplish more tasks, and faster, than individual robots by cooperating and combining their capabilities \cite{karapetyan2017efficient,alonso2017multi}. Large Language Models (LLMs) and Vision–Language Models (VLMs) allow robots to exploit context (e.g., ``repeat your earlier task''), common sense (e.g., ``bring me something to drink''), and general knowledge (e.g., ``from the basket, pick up the fruits rich in vitamin C'') to understand and execute complex instructions.  
This paper presents CoMuRoS (Collaborative Multi-Robot System), a hierarchical architecture for zero-shot, event-driven replanning and recovery in heterogeneous multi-robot teams operating in real-world settings. The design is inspired by human organizations wherein a manager allocates tasks to team members based on their capabilities, and individuals complete their tasks using their skills while proactively reporting progress or problems. In CoMuRoS, a centralized \textit{Task Manager} LLM interprets user intent, breaks down the goal into subtasks, and assigns them to robots according to their capabilities, morphologies, and constraints.  
Tasks are classified into four categories:  
\textbf{a) Independent:} tasks performed in parallel without cooperation (e.g., one robot waters plants while another cooks); 
\textbf{b) Sequential:} tasks requiring successive subtask execution (e.g., a cooking robot prepares food and a serving robot delivers it);  
\textbf{c) Coordinated:} tasks requiring tightly coupled motion (e.g., robots in formation carrying an object together);
\textbf{d) Infeasible:} tasks beyond the system’s capabilities (e.g., asking a differential-drive robot to fly).  
The Task Manager times task initiation based on task type and task completion status reported by robots. Each robot, once assigned a high-level task, uses its local LLM to compose Python code from primitive functions for execution. If a robot detects an event relevant to team's tasks, it notifies the Task Manager, triggering replanning. Replanning allows robots to pause and assist teammates, then resume their own tasks using task-status monitoring, or to assign part of the task to a human when robot capabilities are insufficient. This enables both multi-robot and human–multirobot collaboration.  
The effectiveness of CoMuRoS is demonstrated through hardware and simulation. Hardware experiments show heterogeneous teams recovering autonomously from failures, ignoring irrelevant distractions (e.g., an unrelated object being thrown), and performing coordinated transport tasks where human help emerges naturally when failures occur. Simulation studies highlight the ability to understand human intention and accommodate anytime human interaction, illustrated through hospital and disaster-relief scenarios. A curated dataset of 22 scenarios, each with three tasks and involving around 20 robots across diverse domains, is used to benchmark classification, allocation, planning correctness, executability, and robustness to variation in user instructions and LLM backends.  
The \textit{key contributions} of this work are:
\vspace{-1mm}

\begin{enumerate}
    \item A unified architecture handling independent, sequential, and coordinated multi-robot tasks.  
    \item A curated textual benchmark dataset (22 scenarios, 3 long-horizon tasks each) for evaluating generalization in high-level planning. 
    \item A hierarchical, human-team–inspired design combining centralized task management with decentralized execution, proactive event reporting, and replanning that enables robots to help teammates and resume tasks via status monitoring.  
    \item A modular ROS framework that generalizes to new robots and scenarios through a configuration file (capabilities, scenario-specific rules).  
    \item Support for human–multirobot collaboration where robots request help when failures exceed their capabilities.
    \item Anytime human interaction (feedback, corrections, interruptions, or new commands).  
    \item Hardware experiments and physics-based simulations demonstrating autonomous heterogeneous teams capable of event-driven replanning and recovery.  
\end{enumerate}

\section{RELATED WORK}
\newcommand{\pmark}{$\sim$}     
\begin{table*}[!h]
\centering
\caption{Comparison of LLM-based multi-robot systems.}
\label{tab:comparison}
\resizebox{\textwidth}{!}{
\begin{tabular}{l|c|c|c|c|c|c}
\hline
 & CoELA \cite{zhang2023building} & SMART-LLM \cite{kannan2024smart} & RoCo \cite{mandi2024roco} & COHERENT\cite{liu2024coherent} & DART-LLM \cite{wang2024dart} & CoMuRoS (Ours) \\
\hline
Year & 2024 & 2024 & 2024 & 2025 & 2025 & 2025 \\
Multi-Robot & \ding{51} & \ding{51} & \ding{51} & \ding{51} & \ding{51} & \ding{51} \\
Heterogeneous & \ding{51} & \ding{51} & \ding{51} & \ding{51} & \ding{51} & \ding{51} \\
Framework (Cent/Dec/Hybrid) & Decentralized & Centralized & Decentralized & Hybrid & Hybrid & Hybrid \\
Replanning & \ding{51} & \ding{55} & \ding{51} & \ding{51} & \ding{51} & \ding{51} \\
Event-Driven Replanning (Simulation Demo) & \ding{55} & \ding{55} & \ding{55} & \ding{55} & \ding{55} & \ding{51} \\
Human Collaboration & \ding{55} & \ding{55} & \ding{51} & \ding{55} & \ding{55} & \ding{51} \\
Hardware Demo & \ding{55} & \ding{51} & \ding{51} & \ding{51} & \ding{51} & \ding{51} \\
Event-Driven Replanning (Hardware Demo) & \ding{55} & \ding{55} & \ding{55} & \ding{55} & \ding{55} & \ding{51} \\
Explicit Dependency Modeling & \ding{55} & \ding{55} & \ding{55} & \ding{51} & \ding{51} & Implicit \\
Modular Configuration for Generalization & \ding{55} & \ding{55} & \ding{55} & \ding{55} & \ding{55} & \ding{51} \\
VLM Integration for Perception/Events & \ding{55} & \ding{55} & \ding{55} & \ding{55} & \ding{55} & \ding{51} \\
Opensource ROS2 Framework & \ding{55} & \ding{55} & \ding{55} & \ding{55} & \ding{55} & \ding{51} \\
\hline
\end{tabular}
}
\end{table*}

LLM task planners \cite{huang2022language} and Robot Foundation Models \cite{shah2023lm,firoozi2025foundation} represent two distinct paradigms for robot task planning and execution. Robot Foundation Models or Vision–Language Action Models (VLAs), such as RT-1/RT-2 \cite{brohan2022rt,zitkovich2023rt}, OpenVLA \cite{kim2024openvla}, and GR00T N1 \cite{bjorck2025gr00t}, directly map visual or sensor observations and natural language commands to robot actions. While effective for manipulation and single-robot scenarios, they are not designed for multi-robot systems. Their closed-loop policies are trained on demonstration datasets and fine-tuned for single robots, limiting performance on long-horizon tasks. They also struggle to reason about relevant events (e.g., failures or environment changes) and generalize less effectively than zero-shot LLM planners. In contrast, LLM-based planners use common-sense reasoning, integrate symbolic knowledge of robot capabilities, and deliberate over alternatives, offering stronger robustness in dynamic conditions. VLAs, imitation learning, reinforcement learning (RL) \cite{barto2021reinforcement}, and classical planning–control \cite{spong2008robot} remain valuable at the policy or primitive-action level but require a deliberative planner, such as an LLM, for long-horizon task planning and event-driven replanning.

Early planning frameworks like PDDL \cite{aeronautiques1998pddl} or classical symbolic reasoning \cite{fikes1971strips} work reliably in fully known environments but lack adaptability in open-ended or uncertain domains. Recently, LLM-based planners have emerged as alternatives by combining atomic robot actions into valid sequences in unseen settings. SMART-LLM \cite{kannan2024smart}, for example, demonstrated Python-based planning but lacked ROS2 integration, hierarchical design, or a dialogue interface, and did not test event-driven replanning. COHERENT \cite{liu2024coherent} selects atomic actions from object locations but does not generate Python code, assumes full observability, and has not demonstrated event-driven replanning.

A central challenge in multi-robot settings is balancing centralized and decentralized planning. Frameworks like Decentralized Multi-Agent Systems (DMAS) \cite{zhang2023building} and RoCo \cite{mandi2024roco} explored dialogue-based decentralized systems but suffer from token inefficiency and poor scalability as robot numbers grow. Chen et al. \cite{chen2024scalable} introduced Hybrid Multi-Agent Systems (HMAS-1/2), combining centralized and decentralized reasoning. HMAS-2 showed improved success rate and token efficiency in coordination tasks, though event-driven replanning in dynamic settings was not explored.  

The CoMuRoS framework takes a hierarchical approach where the LLM acts as a high-level planner, using zero-shot reasoning to generate plans, classify tasks, and replan when events occur. Execution is delegated to robot modules handling low-level control via ROS2, classical controllers, or VLA/RL policies. This ensures the LLM focuses on deliberation while robot-level systems ensure reliable execution. Unlike prior work, CoMuRoS supports event-driven replanning, scalable ROS integration, and generalizable coordination across heterogeneous teams.

\label{sec:related_work}

A comparison of existing LLM-based multi-robot planners with CoMuRoS, focusing on framework type, replanning, human collaboration, hardware validation, dependency modeling, modularity, perception/event integration, and ROS2 support is given in Table~\ref{tab:comparison}.  

Explicit dependency modeling, as in DART-LLM’s \cite{wang2024dart} Directed Acyclic Graph (DAG) or COHERENT’s \cite{liu2024coherent} Proposal–Execution–Feedback–Adjustment (PEFA), enforces deterministic ordering and correctness in well-defined workflows. However, these methods are rigid when unexpected events or new agents appear, require heavy engineering to encode preconditions and effects, and usually assume near-complete observability. In contrast, CoMuRoS resolves dependencies implicitly through context-driven reasoning, task classification, task-status monitoring, and event feedback. Robots proactively report completions and interruptions, enabling the Task Manager to infer sequential dependencies, launch parallel tasks, or trigger replanning. This trades strict guarantees for adaptability, supporting event-driven replanning, seamless human interaction, generalization, and emergent multi-robot and human–multirobot collaboration.  

While RoCo \cite{mandi2024roco}, COHERENT \cite{liu2024coherent}, DART-LLM \cite{wang2024dart}, and CoELA \cite{zhang2023building} include some form of replanning, all remain limited and lack runtime recovery in hardware. RoCo uses a validation loop where infeasible subtasks (e.g., failing IK or collisions) are corrected before execution, but no mid-execution replanning occurs. COHERENT’s PEFA loop iteratively refines plans using robot feedback from completed tasks rather than unexpected events. DART-LLM represents tasks as a DAG, deferring or rescheduling subtasks when prerequisites fail, but it does not generate entirely new plans or reassign tasks to other robots. CoELA emphasizes decentralized planning and communication efficiency, but lacks event-driven replanning. By contrast, CoMuRoS uniquely demonstrates event-driven replanning on physical robots: failures such as object drops or human intent changes trigger the Task Manager to generate a new plan, reallocate tasks (including to other robots or humans), and continue execution. This dynamic recovery enables emergent cooperation in both multi-robot and human–robot settings, absent in prior work.
\vspace{-1mm}
\section{METHODOLOGY}
\begin{figure*}[!t]
    \centering
        \includegraphics[width=0.7\linewidth]{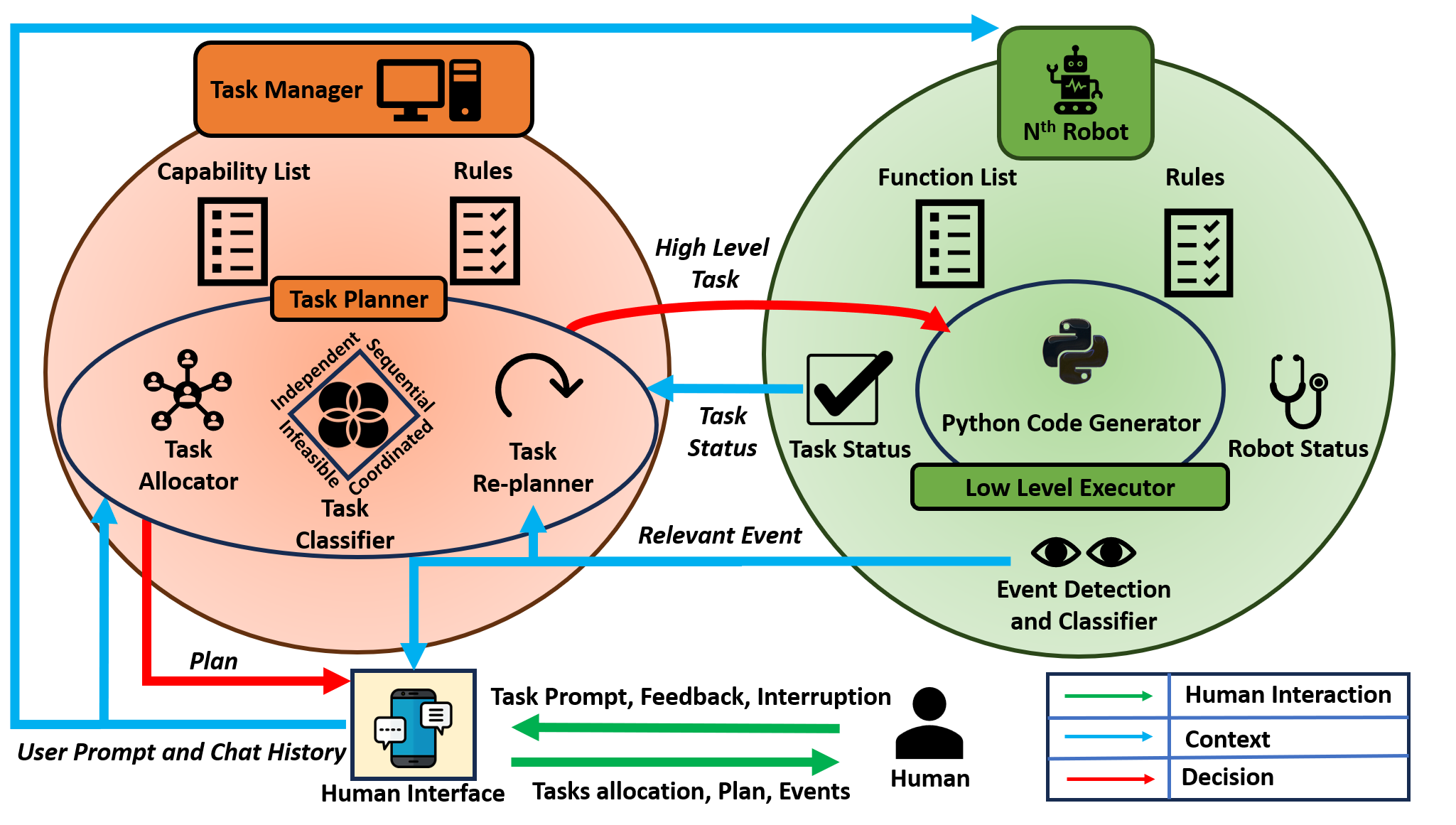}
    \caption{Architectural Diagram of CoMuRoS}
    \label{fig:arch_diagram}
\end{figure*}
\begin{figure*}[!t]
    \centering
    \setlength{\belowcaptionskip}{-15pt}
    
    \includegraphics[width=0.9\linewidth, trim=12 0 8 9, clip]{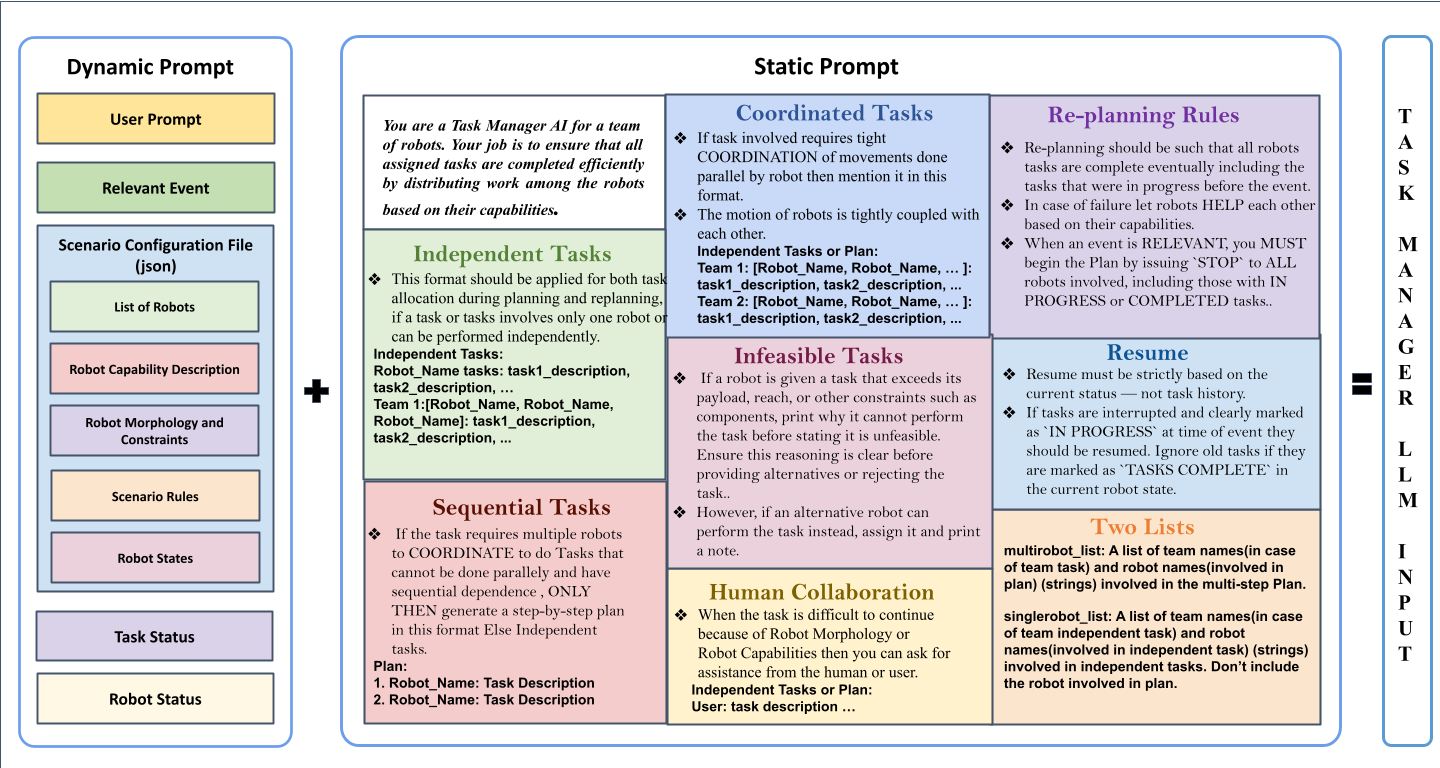}

    \caption{Task Manager Prompt}
    \label{fig:prompt_diagram}
\end{figure*}
\label{sec:methodology}
CoMuRoS is a hierarchical architecture inspired by human organizational structures. In this analogy, the user plays the role of a CEO or chairman, providing high-level goals in natural language to a manager. The manager, with knowledge of the portfolios and capabilities of each agent, allocates tasks accordingly. Each team member then decides how to combine their available skills for execution and proactively reports back to the manager in case of any significant events.  

The hierarchical architecture (Fig.~\ref{fig:arch_diagram}) represents a hybrid of centralized and decentralized paradigms. A centralized Task Manager LLM is responsible for high-level planning, while each robot maintains its own LLM for low-level execution, code generation, and event classification.
\vspace{-1mm}\vspace{-1mm}\vspace{-1mm}
\subsection{Task Manager LLM}
The Task Manager LLM has three primary responsibilities: task classification, task allocation, and task replanning in response to relevant events.  

The Task Manager classifies user-provided tasks as \textit{sequential}, \textit{independent}, \textit{coordinated}, or \textit{infeasible}. It then allocates tasks to robots based on their capabilities, morphologies, and constraints. When a robot or human communicates an update that influences execution, the Task Manager replans to ensure that all tasks are completed successfully.  

Replanning leverages the context of chat history, which includes prior tasks, robot status, events, and task statuses. Prompt engineering is employed to exploit the common-sense reasoning and context awareness of LLMs for understanding user commands, classifying tasks and events, and allocating tasks according to robot capabilities. Temperature=0.5 for all experiments in this paper.

As shown in Fig. \ref{fig:prompt_diagram}, the Task Manager prompt is divided into static and dynamic components. The static prompt encodes rules, definitions, formats, and the procedural logic required for planning and replanning. It forms the “brain” of the Task Manager by containing memory of task definitions, allocation strategies, and replanning procedures. The dynamic prompt contains scenario-specific information, including user commands, chat history, robot status, task status, and detected events. The scenario configuration file specifies the robots’ capabilities along with scenario-specific rules.
Task status can take three values: \texttt{COMPLETED}, \texttt{IN PROGRESS}, or \texttt{INTERRUPTED}, as reported by robots. During replanning, only tasks not marked \texttt{COMPLETED} are reconsidered. Robot status is also important in sequential tasks, since some actions are feasible only in specific states (e.g., a robotic arm with limited reach can place an object on a quadruped only if the quadruped is sitting).

Human–robot collaboration is also supported. If humans are included in the dynamic prompt, the Task Manager may assign tasks to humans when no robot can execute them.  
\vspace{-1mm}
\subsection{Robot Brain}
The architecture is robot-agnostic and supports heterogeneous platforms. Each robot receives high-level tasks in natural language from the Task Manager, which ensures feasibility given the robot’s constraints.  

As shown in Fig.~\ref{fig:arch_diagram}, each robot maintains a list of execution functions with associated descriptions (inputs, outputs, and use cases). These functions may correspond to reinforcement learning policies, imitation learning models, Python code, or ROS nodes. The robot’s LLM decomposes the high-level command and combines relevant functions to generate Python code for execution, thereby producing a ready-to-run low-level plan.  

Robots are agnostic to whether tasks are sequential or independent. Each robot processes one high-level command at a time and, upon completion, updates the task status to \texttt{COMPLETED}. Execution functions must include mechanisms to update both task and robot status, either through sensor feedback or open-loop logic. Robot status may include descriptors such as “sitting,” “standing,” or coordinates such as $(4,0)$.
Robots are also responsible for proactively detecting and reporting events in their environment using onboard sensors or external inputs (e.g., CCTV). Event detection operates in parallel with task execution. An LLM is employed to classify detected events as relevant (affecting task progress) or irrelevant (e.g., “a dog is barking” when robots are tasked with indoor cleaning). Relevant events are reported to the Task Manager, which triggers replanning. Event detection methods may include image processing, action recognition, or vision–language models.  
\vspace{-1mm}\vspace{-1mm}
\subsection{Human Interface}
\begin{figure}[!t]
    \centering   \setlength{\belowcaptionskip}{-15pt}\includegraphics[width=\linewidth]{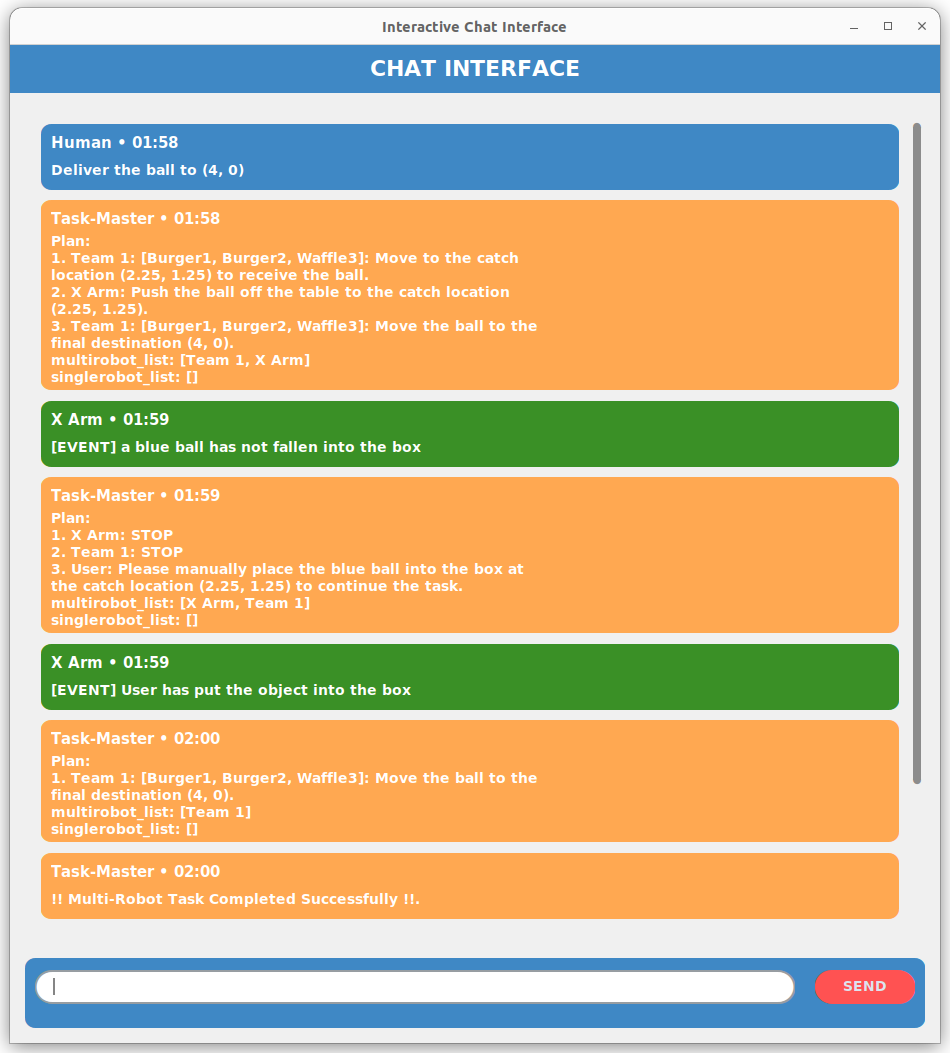}
    \caption{Chat Interface for human robot collaboration.}
\label{fig:human_interface}
\end{figure}
A chat-based interface enables continuous natural language communication between humans and the multi-robot system, as shown in Fig.~\ref{fig:human_interface}. Through this interface, the user can provide information, give new commands, interrupt ongoing tasks, or change intention anytime during execution. The dialogue is color-coded with user commands in blue, task allocations from the Task Manager in orange, and robot-detected events in green, all displayed in temporal order. The chat history is preserved so that instructions like ``repeat the earlier task’’ or ``pick it up again’’ can be understood in context, and so that the Task Manager has memory of past tasks and events for replanning. This also makes replanning transparent, since the triggering event and the updated plan appear together in the interface. User input is given by typing in the interface, but can be extended to speech, and multilingual interaction is supported depending on the LLM used.
\captionsetup[subfigure]{belowskip=0pt,aboveskip=1pt}
\begin{figure*}[!h]
    \centering
    \setlength{\belowcaptionskip}{-15pt}
    \captionsetup[subfigure]{font=footnotesize,aboveskip=0pt,belowskip=0pt}

    \begin{subfigure}[t]{0.31\textwidth}
        \includegraphics[width=\linewidth]{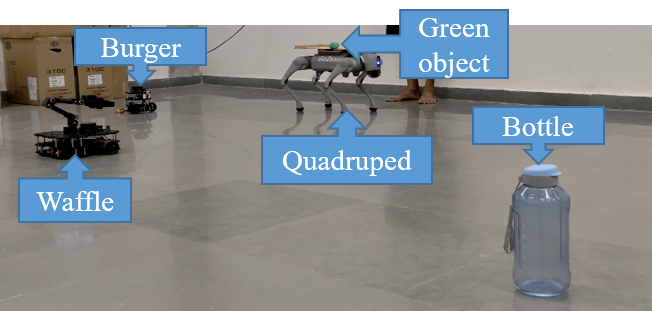}
        \caption{Initial scene overview.}
        \label{fig:hardware_replan_A}
    \end{subfigure}\hfill%
    \begin{subfigure}[t]{0.31\textwidth}
        \includegraphics[width=\linewidth]{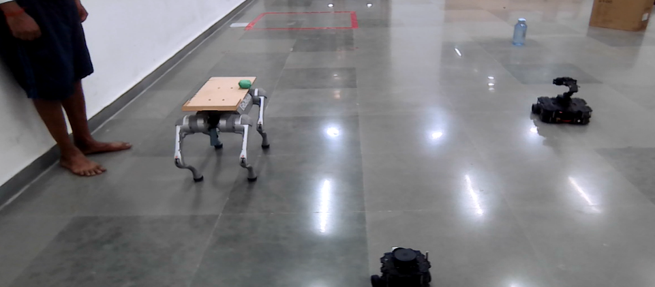}
        \caption{Waffle moves toward the bottle; quadruped heads to its goal with the green object.}
        \label{fig:hardware_replan_B}
    \end{subfigure}\hfill%
    \begin{subfigure}[t]{0.31\textwidth}
        \includegraphics[width=\linewidth]{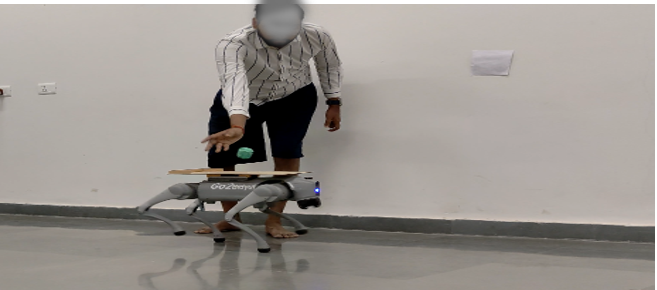}
        \caption{Human throws the green object off the quadruped’s back.}
        \label{fig:hardware_replan_C}
    \end{subfigure}

    \par\medskip

    \begin{subfigure}[t]{0.31\textwidth}
        \includegraphics[width=\linewidth]{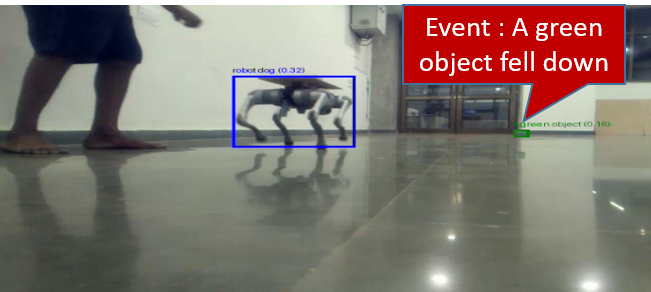}
        \caption{Burger detects the fall and raises an event; Task Manager replans.}
        \label{fig:hardware_replan_D}
    \end{subfigure}\hfill%
    \begin{subfigure}[t]{0.31\textwidth}
        \includegraphics[width=\linewidth]{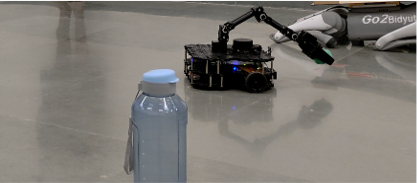}
        \caption{Waffle picks up the green object.}
        \label{fig:hardware_replan_E}
    \end{subfigure}\hfill%
    \begin{subfigure}[t]{0.31\textwidth}
        \includegraphics[width=\linewidth]{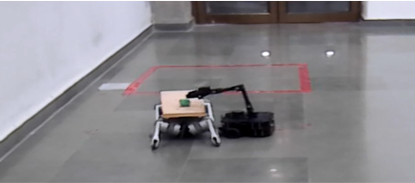}
        \caption{Quadruped sits; Waffle places the object back.}
        \label{fig:hardware_replan_F}
    \end{subfigure}

    \par\medskip

    \begin{subfigure}[t]{0.31\textwidth}
        \includegraphics[width=\linewidth]{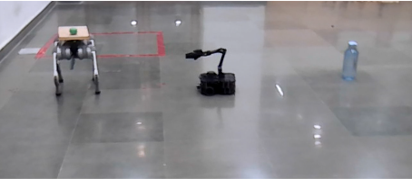}
        \caption{Quadruped stands.}
        \label{fig:hardware_replan_G}
    \end{subfigure}\hfill%
    \begin{subfigure}[t]{0.31\textwidth}
        \includegraphics[width=\linewidth]{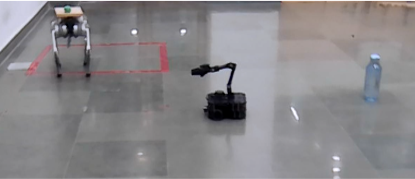}
        \caption{Quadruped goes to the goal.}
        \label{fig:hardware_replan_H}
    \end{subfigure}\hfill%
    \begin{subfigure}[t]{0.31\textwidth}
        \includegraphics[width=\linewidth]{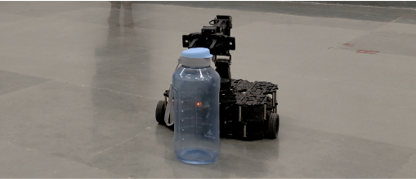}
        \caption{Waffle resumes approaching the bottle.}
        \label{fig:hardware_replan_I}
    \end{subfigure}

    \caption{Demonstration of event-driven replanning, incomplete task resumption, and emergence of cooperation where robots assist one another.}
    \label{fig:hardware_event_replan}
\end{figure*}
\vspace{-1mm}
\subsection{Evaluation Metrics}
We constructed a textual dataset of 22 diverse scenarios, each with three tasks, to evaluate CoMuRoS. Each scenario includes a configuration file with robot lists, capabilities, and scenario-specific rules, along with three human input prompts(tasks). Textual dataset is evaluated under the following metrics: 
\textbf{Task Allocation (TA)} equals 1 if tasks are correctly assigned according to robot's capabilities and user task, otherwise 0; 
\textbf{Task Classification (TC)} equals 1 if the classification is feasible given the allocation, otherwise 0 (some tasks may be performed both sequentially and independently); 
\textbf{Intersection-over-Union (IoU)} is the standard metric measuring overlap between the Task Manager’s plan and ground truth (0–1); 
\textbf{Executability (Exec)} is the percentage of feasible allocations given robot capabilities and constraints (normalized to 0–1); and 
\textbf{Correctness} equals 1 if an expert evaluator judges the plan to succeed, else 0.
\vspace{-1mm}
\section{Experiments and Results}
\vspace{-1mm}
\subsection{Hardware}

\begin{figure}[h]
    \centering
    \setlength{\belowcaptionskip}{-25pt}

    \begin{subfigure}[t]{0.48\linewidth}
        \captionsetup{font=footnotesize} 
        \includegraphics[width=\linewidth]{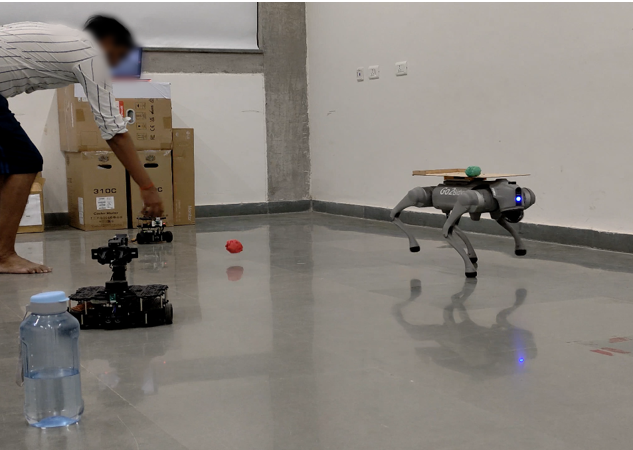}
        \caption{Human throws a red object in front of Burger robot.}
        \label{fig:hardware_event_C}
    \end{subfigure}\hfill
    \begin{subfigure}[t]{0.48\linewidth}
        \captionsetup{font=footnotesize}
        \includegraphics[width=\linewidth]{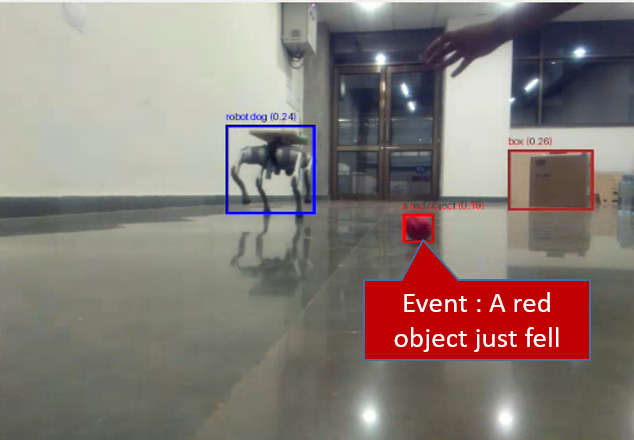}
        \caption{Burger detects the red object fall event but ignores it as it is irrelevant to the task.}
        \label{fig:hardware_event_D}
    \end{subfigure}

    \begin{subfigure}[t]{0.48\linewidth}
        \captionsetup{font=footnotesize}
        \includegraphics[width=\linewidth]{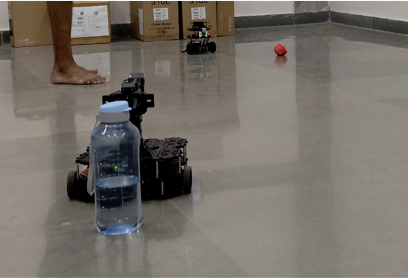}
        \caption{Waffle reaches the bottle.}
        \label{fig:hardware_event_E}
    \end{subfigure}\hfill
    \begin{subfigure}[t]{0.48\linewidth}
        \captionsetup{font=footnotesize}
        \includegraphics[width=\linewidth]{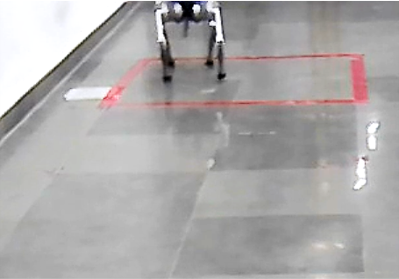}
        \caption{Quadruped reaches its goal location.}
        \label{fig:hardware_event_F}
    \end{subfigure}

    \caption{Demonstration of ignoring irrelevant events.} 
    \label{fig:hardware_event_interrupted}
\end{figure}
To demonstrate CoMuRoS capabilities in task allocation, event detection, classification, and event-driven replanning with onboard sensors, two sets of hardware experiments were conducted: (1) Multi-robot collaboration with replanning, and (2) Human–formation collaboration.\subsubsection{Multi-robot Collaboration using Event-Driven Replanning}
We deploy a Unitree Go2 quadruped, a Turtlebot Burger, and a Turtlebot Waffle-pi with an OpenManipulator-X arm (Waffle). The objective is to test whether independently tasked robots can recover collaboratively when a disruptive event occurs. The user instructs Waffle to approach a bottle (local LLM composes executable Python from its function library, e.g., find(object) and reach(object)), and the Go2 to carry a green object to $(3,0)$.

Midway, the green object is deliberately knocked off the Go2 (Fig.~\ref{fig:hardware_replan_C}). CoMuRoS performs (i) online event detection and classifies it as relevant, (ii) triggers event-driven replanning, and (iii) reallocates sequential roles while respecting physical constraints. From the scenario configuration, the Task Manager infers that Waffle’s manipulator lacks reach to place the object on a standing quadruped and therefore instructs the Go2 to sit before the handover (Fig.~\ref{fig:hardware_replan_F}). After Waffle completes the pick-and-place, Go2 resumes delivery (Fig.~\ref{fig:hardware_replan_H}) and Waffle automatically resumes its original task(Fig.~\ref{fig:hardware_replan_I}). Since its status was marked \texttt{INTERRUPTED} in the original plan, the Task Manager explicitly included it in the replanned sequence, ensuring completion of all assigned tasks. Across 10 trials (using GPT 4o), the success rate was \textbf{9/10}; failure arose from incorrect replanning sequence.

A separate experiment demonstrates filtering of irrelevant events. When a red object is thrown in front of Burger (Fig.~\ref{fig:hardware_event_C}), the robot detects the object (Fig.~\ref{fig:hardware_event_D}) but classifies it as irrelevant, so no event is communicated to the Task Manager. Both Waffle and Go2 complete their assigned tasks without interruption (Figs.~\ref{fig:hardware_event_E}–\ref{fig:hardware_event_F}). This shows how decentralized event classification reduces unnecessary processing load on the Task Manager while keeping the system focused on relevant events only.
\subsubsection{Human-Formation Collaboration}
\begin{figure}[!t]
    \centering
    \setlength{\belowcaptionskip}{-15pt}
    
    \captionsetup[subfigure]{font=footnotesize,aboveskip=0pt,belowskip=0pt}

    \begin{subfigure}[t]{0.48\linewidth}
        \includegraphics[width=\linewidth, trim=0 70 0 0, clip]{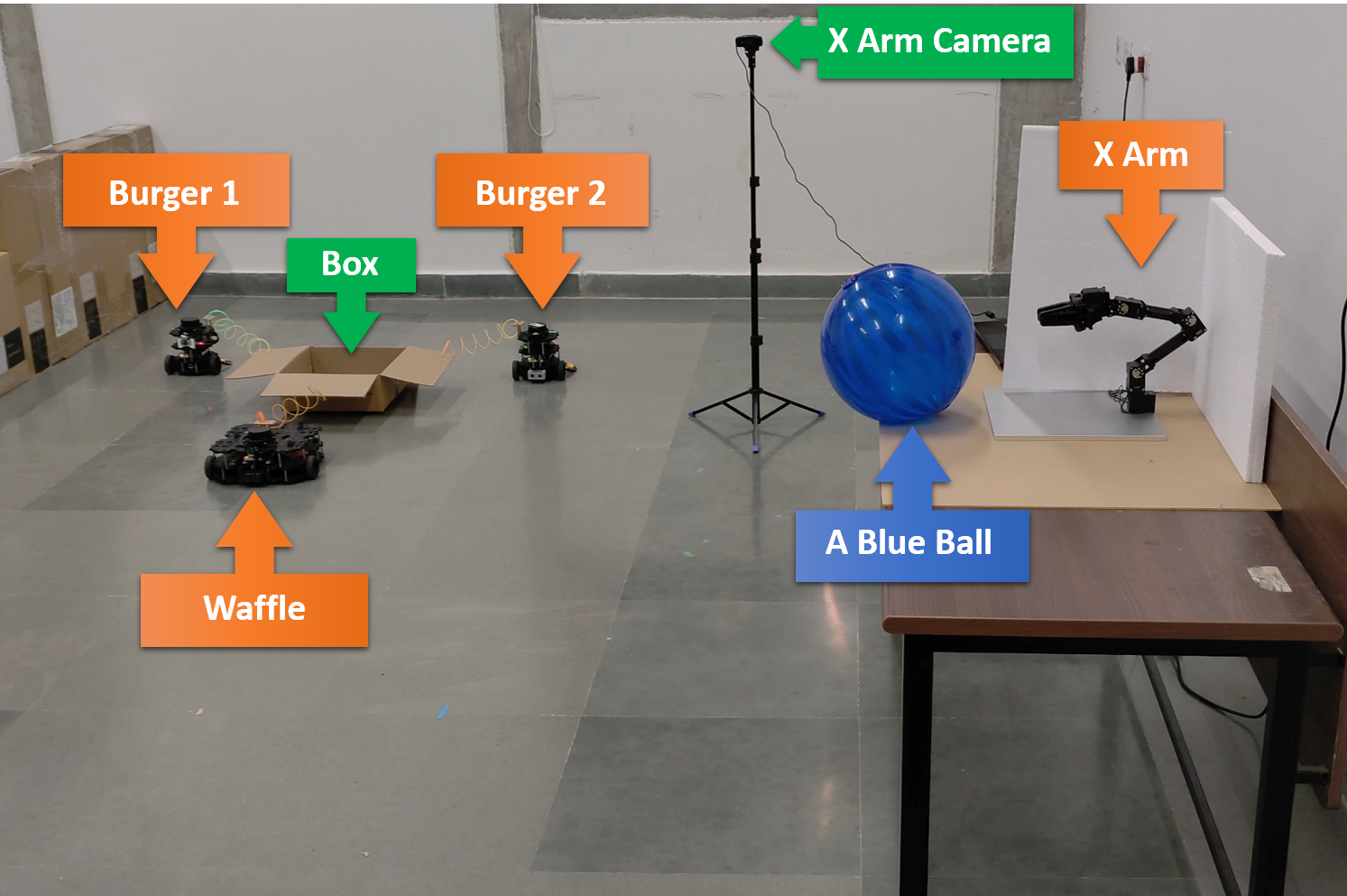}
        \caption{Initial scene overview.}
        \label{fig:sequence_A}
    \end{subfigure}\hfill
    \begin{subfigure}[t]{0.48\linewidth}
        \includegraphics[width=\linewidth]{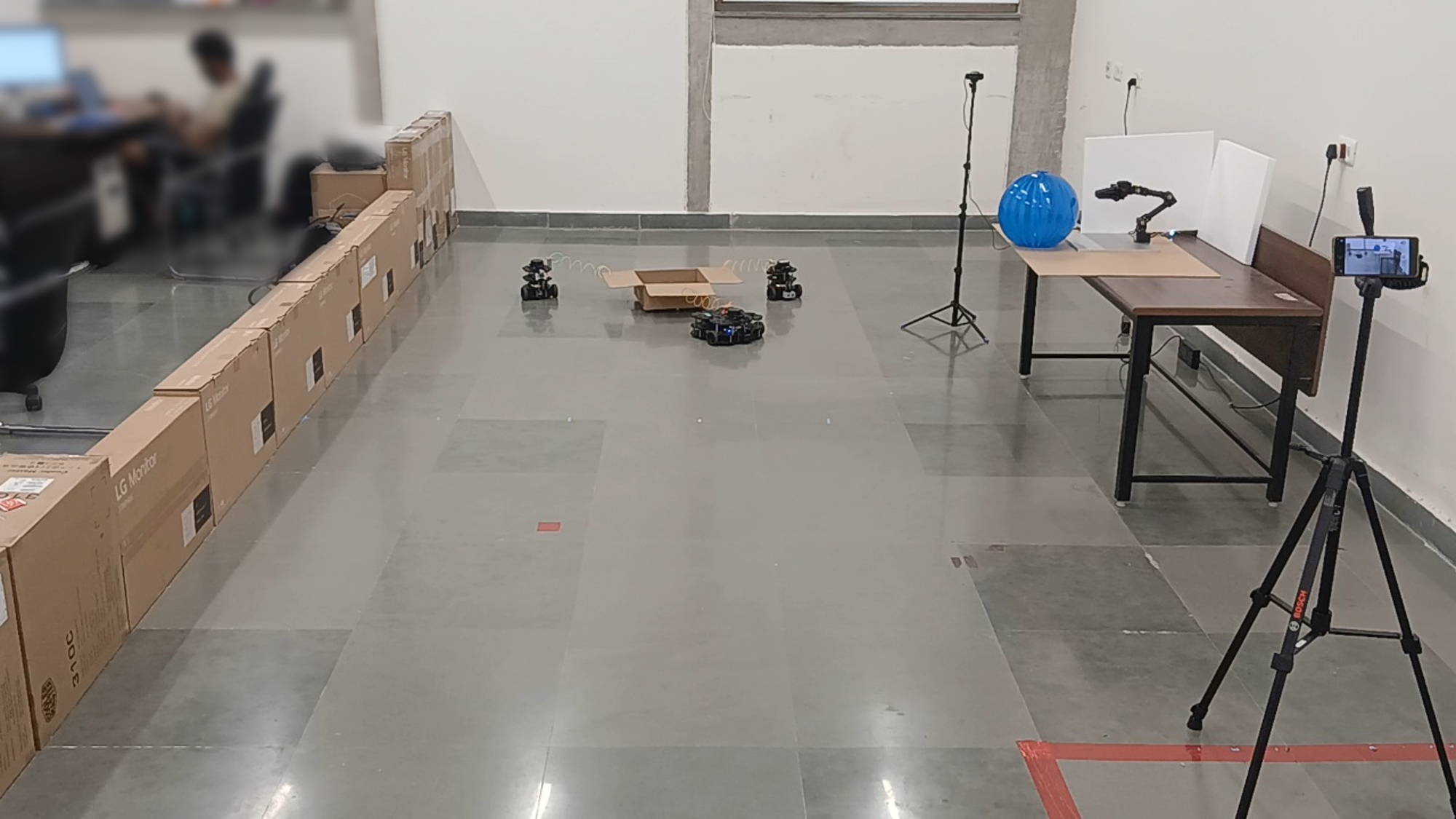}
        \caption{Formation moving to catch point.}
        \label{fig:sequence_B}
    \end{subfigure}

    \begin{subfigure}[t]{0.48\linewidth}
        \includegraphics[width=\linewidth]{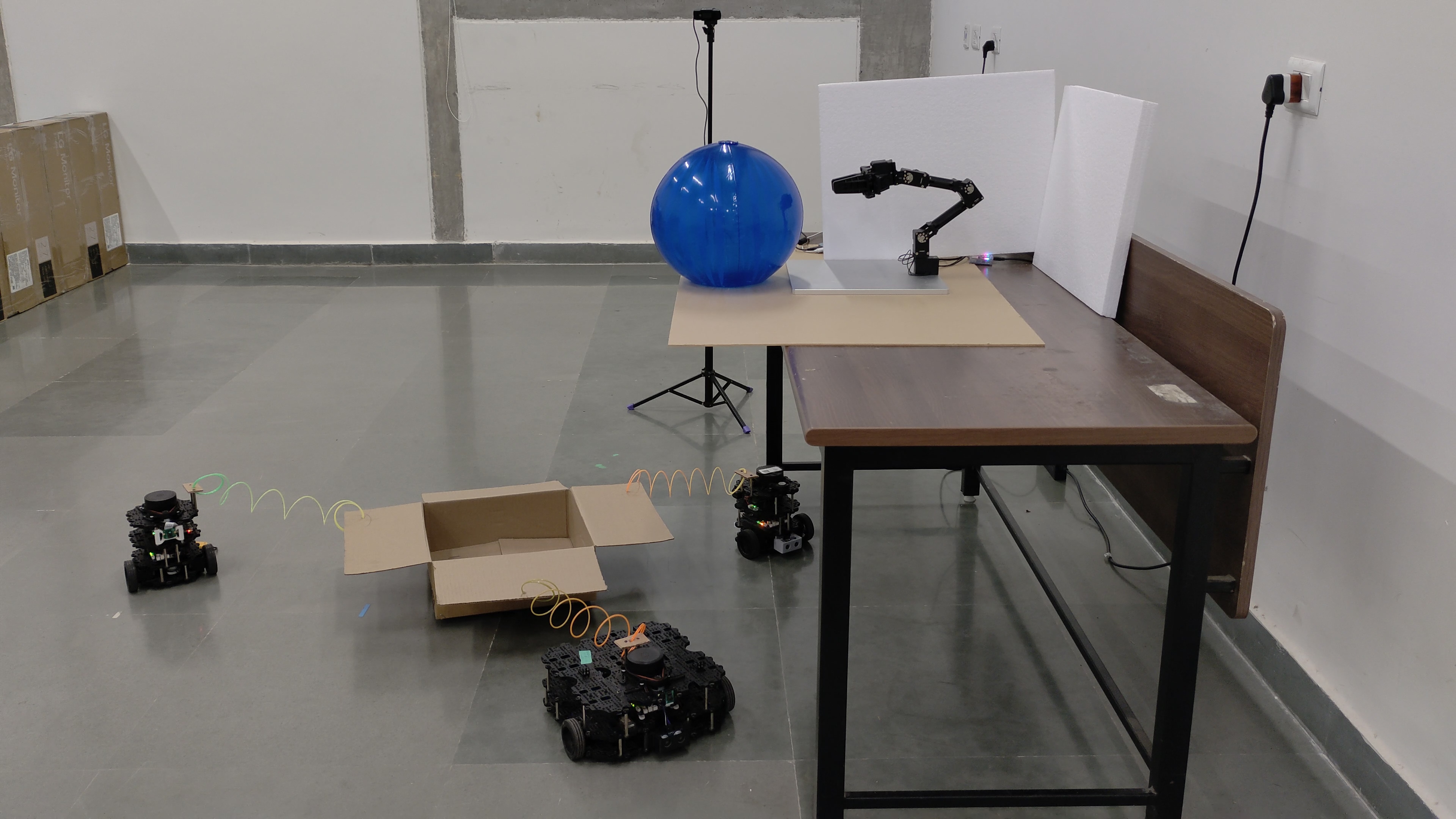}
        \caption{X-arm pushing ball.}
        \label{fig:sequence_C}
    \end{subfigure}\hfill
    \begin{subfigure}[t]{0.48\linewidth}
        \includegraphics[width=\linewidth]{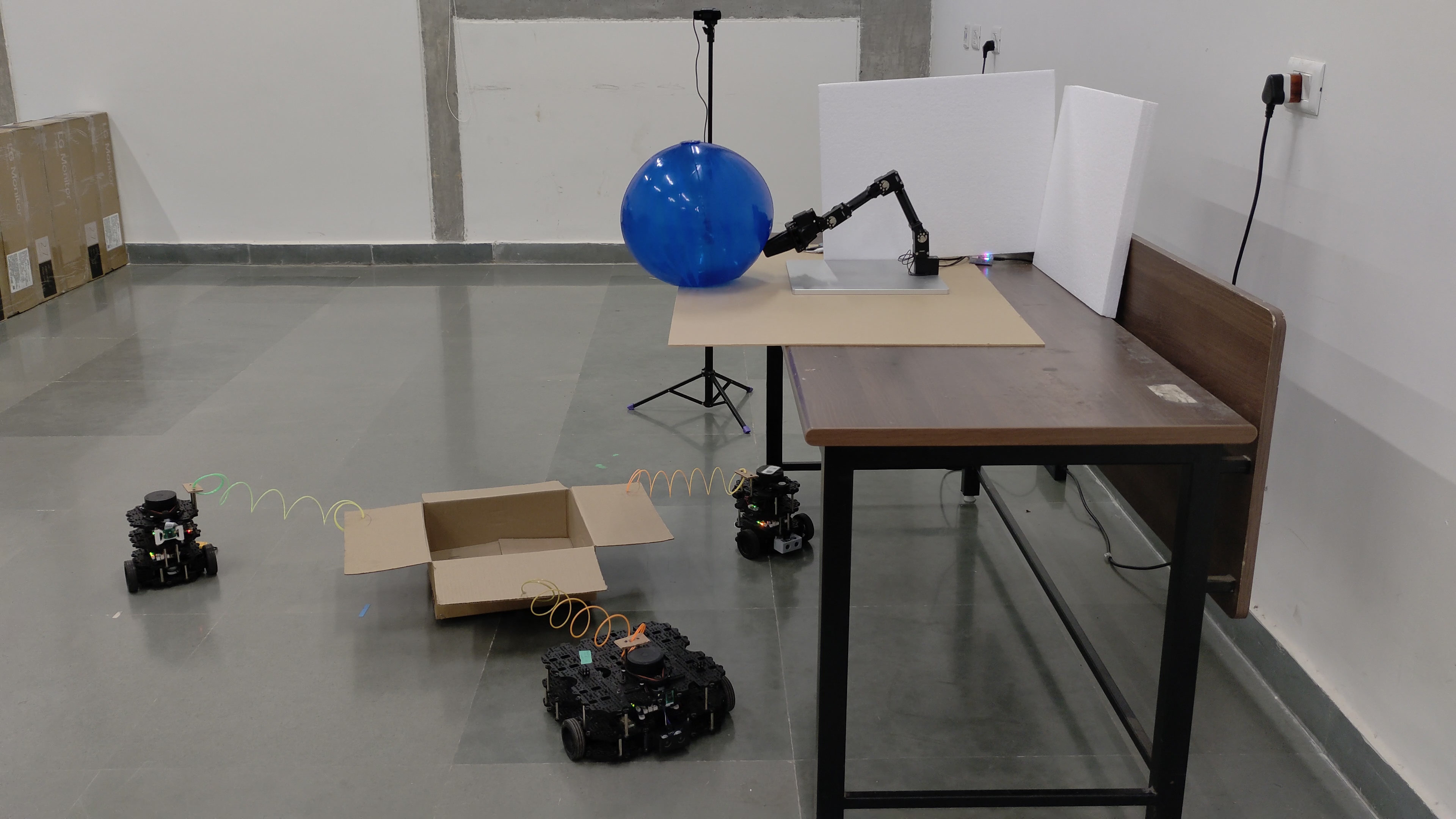}
        \caption{Ball pushed into box.}
        \label{fig:sequence_D}
    \end{subfigure}

    \begin{subfigure}[t]{0.48\linewidth}
        \includegraphics[width=\linewidth]{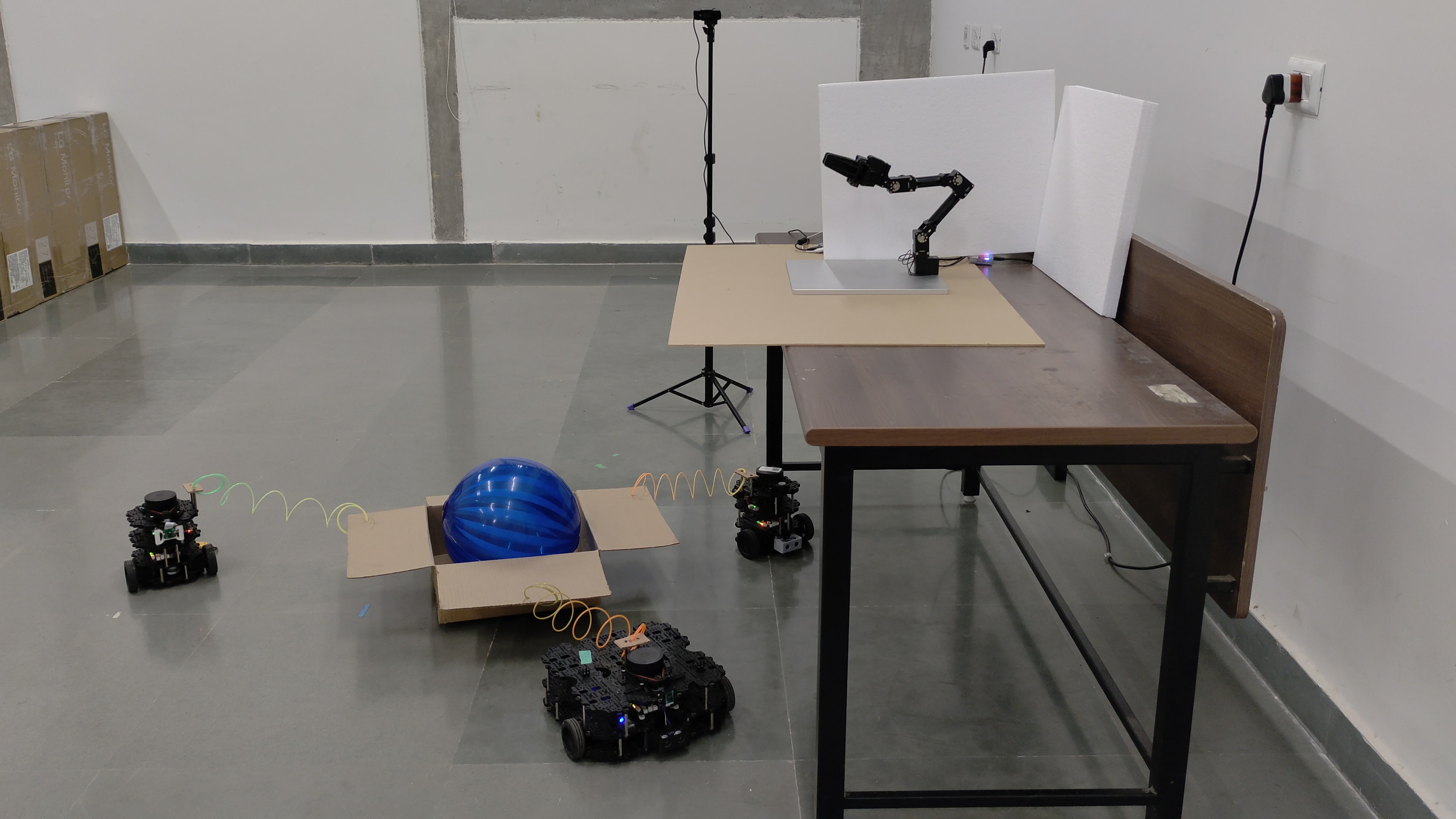}
        \caption{Ball inside box.}
        \label{fig:sequence_E}
    \end{subfigure}\hfill
    \begin{subfigure}[t]{0.48\linewidth}
        \includegraphics[width=\linewidth]{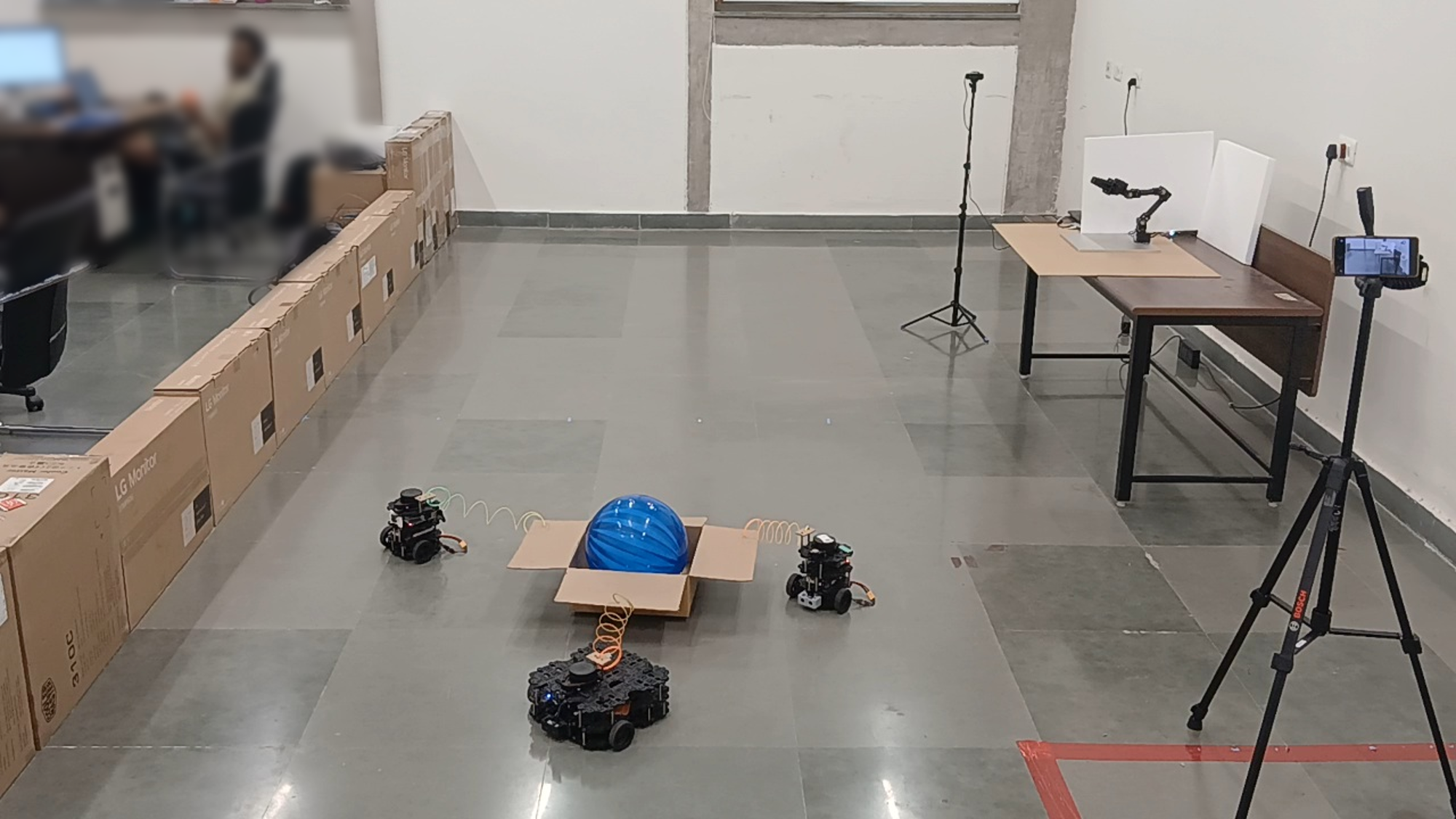}
        \caption{Robots deliver ball to (4,0).}
        \label{fig:sequence_F}
    \end{subfigure}

    \caption{Demonstration of CoMuRoS performing a coordinated task: the X-arm and a mobile robot formation collaborate to transfer the ball from the table to the goal location (4,0).}
    \label{fig:sequence}
\end{figure}
\begin{figure*}[!h]   
    \centering
    \setlength{\belowcaptionskip}{-15pt}

    \begin{subfigure}[t]{0.28\linewidth}
        \includegraphics[width=\linewidth]{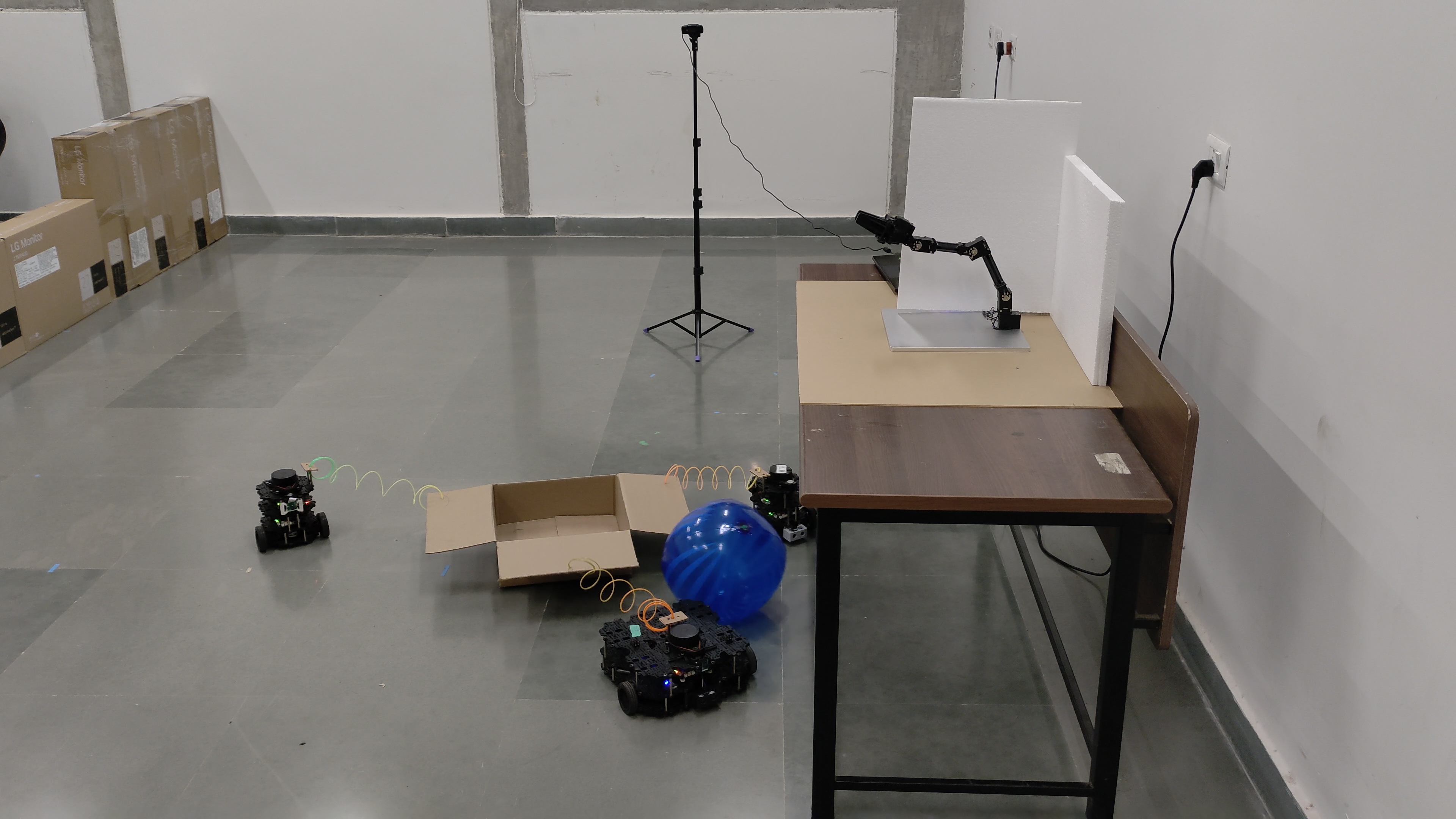}
        \caption{Ball falls outside the box instead of inside.}
        \label{fig:human_helpo_D}
    \end{subfigure}\hfill
    \begin{subfigure}[t]{0.28\linewidth}
        \includegraphics[width=\linewidth]{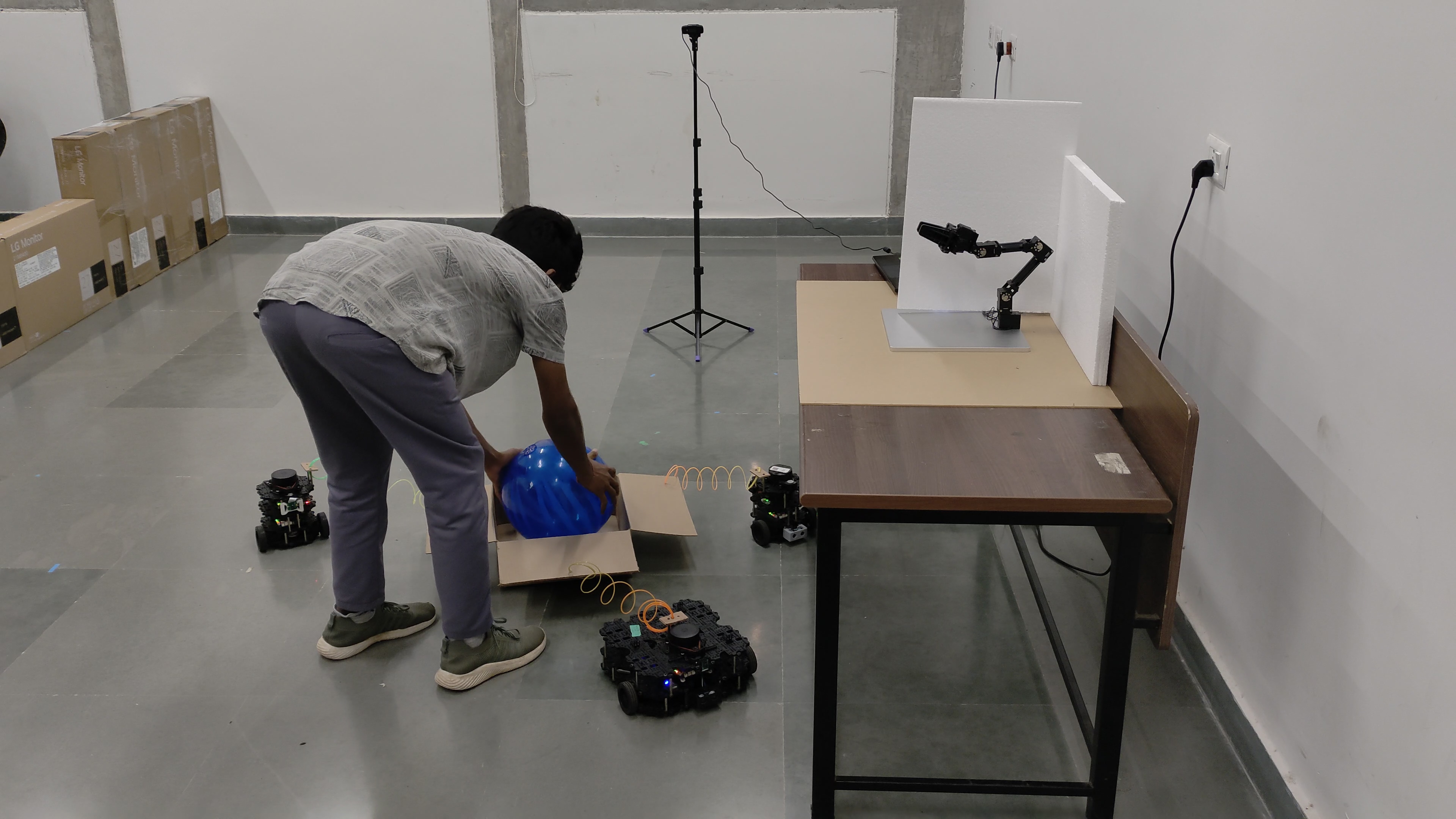}
        \caption{On request from the Task Manager, the human puts the ball inside the box.}
        \label{fig:human_helpo_E}
    \end{subfigure}\hfill
    \begin{subfigure}[t]{0.28\linewidth}
        \includegraphics[width=\linewidth]{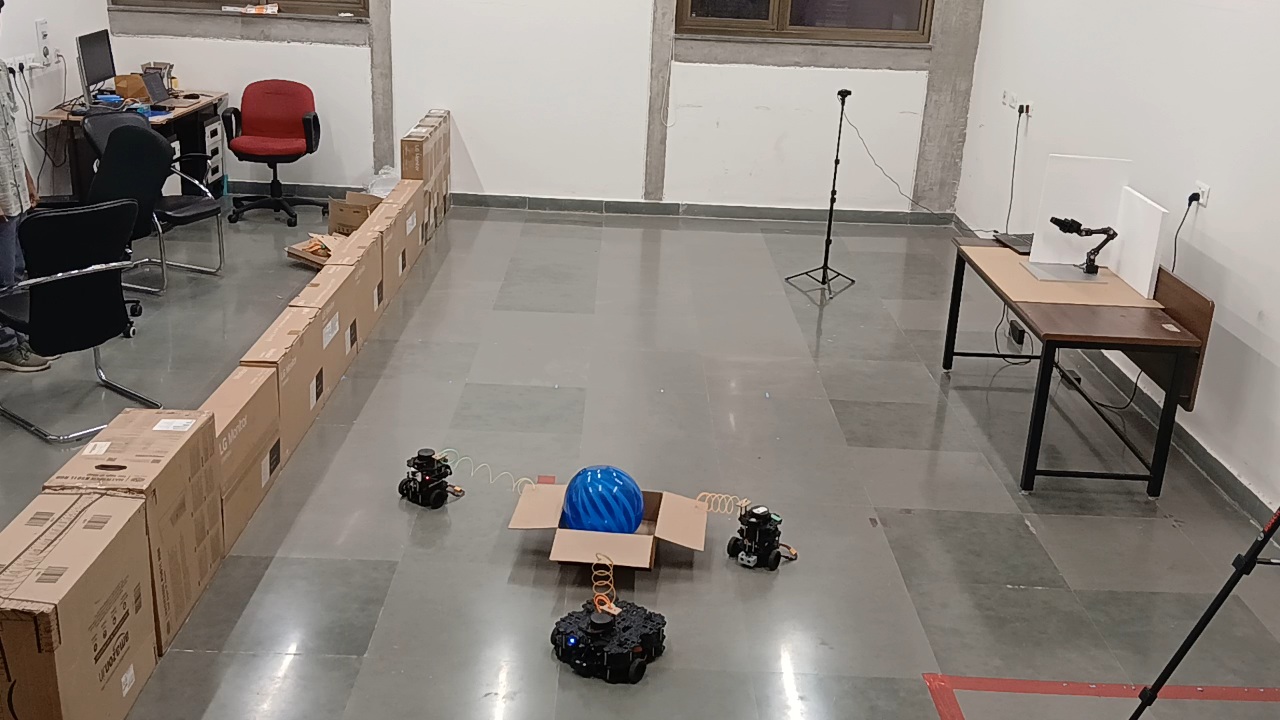}
        \caption{Formation carries the ball towards the goal.}
        \label{fig:human_helpo_F}
    \end{subfigure}

    \caption{Demonstration of human–multi-robot collaboration resulting from failure.}
    \label{fig:human_help}
\end{figure*}

Fig. \ref{fig:sequence}a shows the scene overview of this experiment. A formation of three mobile robots (One Waffle and two Burger robots) can carry a box. There is a table with a blue ball and an OpenManipulator-X arm (X-arm). The user instructs the multi-robot system to "Deliver the blue ball to (4,0)". The scenario configuration file has the description of the scene and catch point coordinates where the formation can collect the ball from the X-arm. The Task Manager classifies the task as sequential and coordinated, generating a plan shown in the chat interface Fig. \ref{fig:human_interface}. According to this plan and as shown in Fig. \ref{fig:sequence_B} to \ref{fig:sequence_E} formation of mobile robots carries the box to the catch point, and the X-arm pushes the ball off the table. The X-arm camera detects whether the ball has fallen into the box successfully. If the ball falls into the box, the X-arms task status is marked \texttt{COMPLETED} and the formation delivers the box to the goal position successfully. This experiment had a success rate is \textbf{8/8}. Fig. \ref{fig:human_help} shows what happens if the X-arm task fails, i.e after pushing, the ball falls outside the box. The X-arm's camera detects the event, notifies it to the task manager and to the human chat interface, as shown in green in Fig. \ref{fig:human_interface}. This event, being relevant to the task, triggers replanning. Since none of the robots has the ability to keep the ball back inside the box, the task manager assigns the task to the user as shown in Fig. \ref{fig:human_interface}.
User places the ball into the box and then the formation carries the ball to goal position successfully, as shown in Fig. \ref{fig:human_help}. The events of the ball falling inside the box or outside, or the placement of the ball inside the box, were detected using a combination of OwlViT-based object detection and GPT 4.1 for Visual Question Answering (VQA). This experiment shows the ability of CoMuRoS architecture to enable human-robot collaboration, the ability to identify if a task is beyond robots' capabilities and in that case, ask for help. This experiment has a success rate of \textbf{5/5}. In order for the ball to fall outside the box, the table position was slightly modified compared to the previous experiment. Human collaboration can significantly improve the robustness of task completions in real settings.

\vspace{-1mm}
\subsection{Simulation}
\begin{figure}[h]
    \centering
    \setlength{\belowcaptionskip}{-14pt}
    \captionsetup[subfigure]{font=footnotesize,aboveskip=0pt,belowskip=0pt}

    \begin{subfigure}[t]{0.48\linewidth}
        \includegraphics[width=\linewidth]{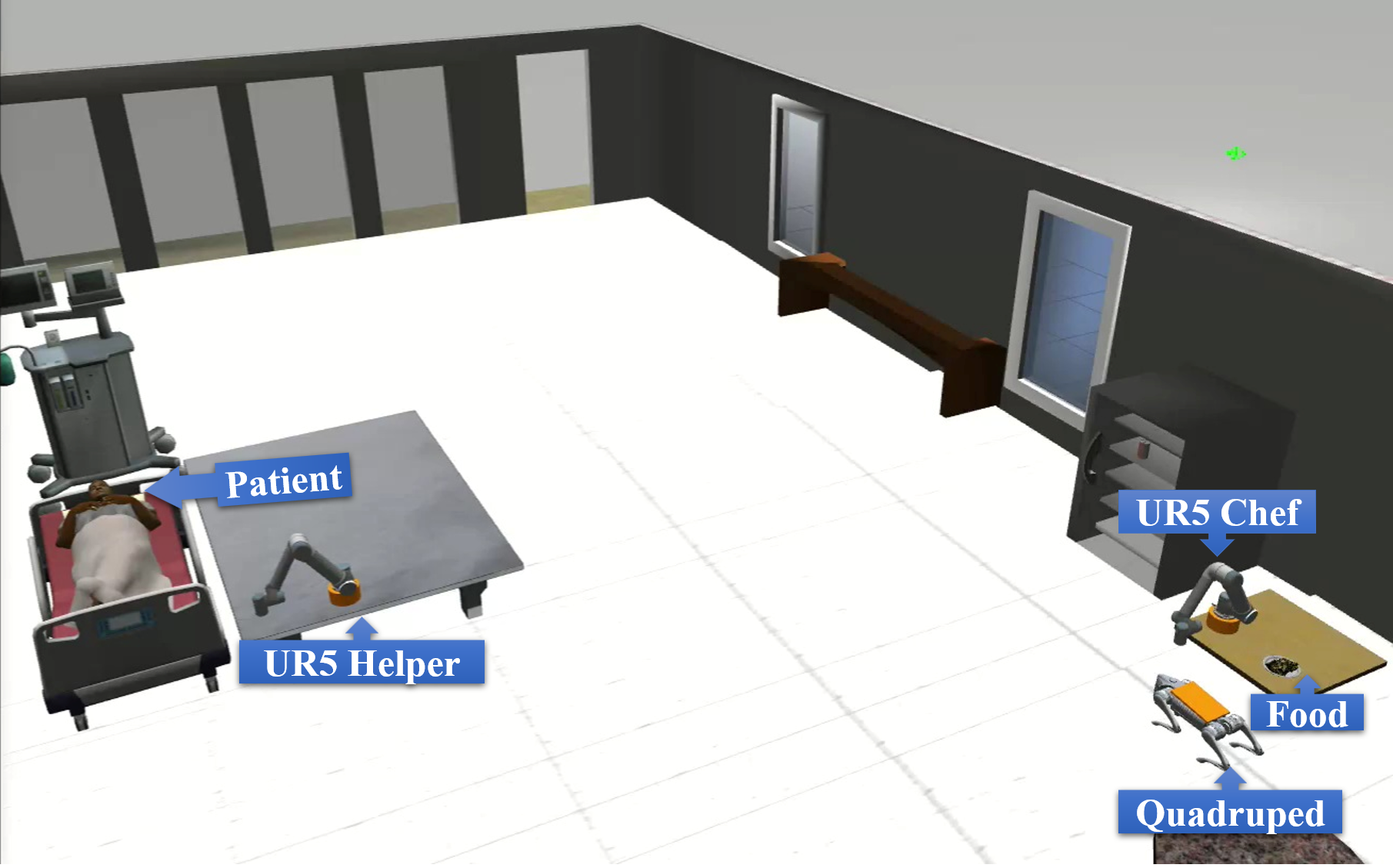}
        \caption{Initial scene overview.}
        \label{fig:hospital_A}
    \end{subfigure}\hfill
    \begin{subfigure}[t]{0.48\linewidth}
        \includegraphics[width=\linewidth]{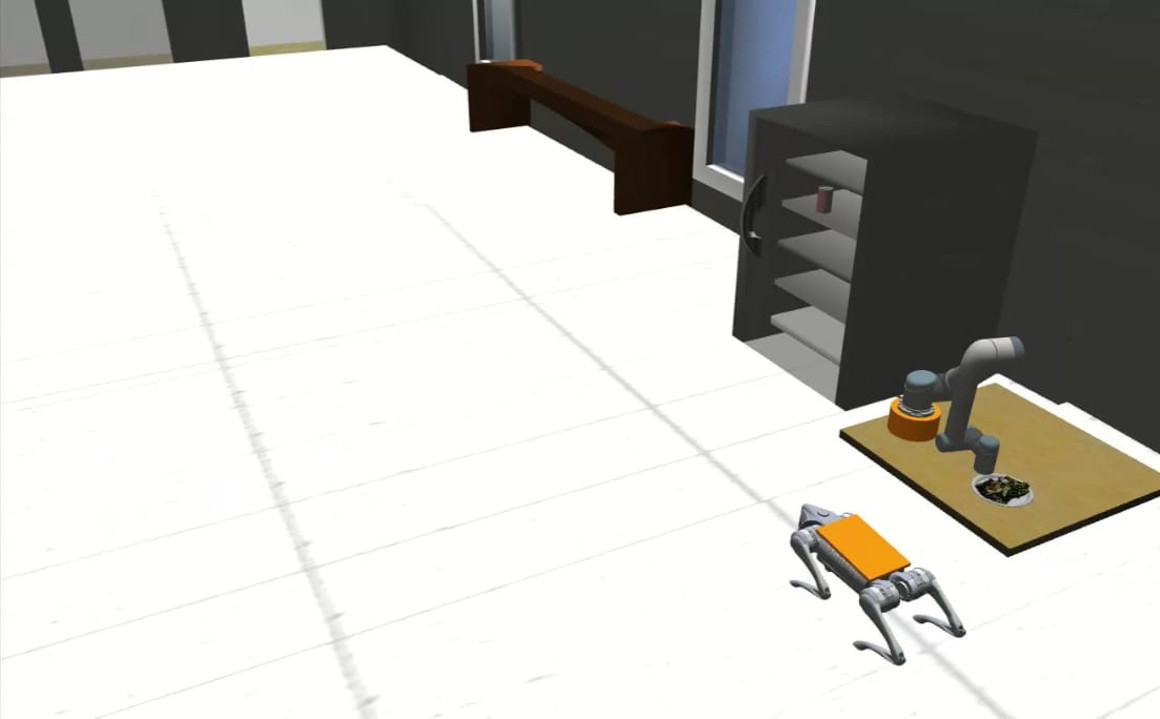}
        \caption{Patient says: ``I am hungry.''}
        \label{fig:hospital_B}
    \end{subfigure}

    \begin{subfigure}[t]{0.48\linewidth}
        \includegraphics[width=\linewidth]{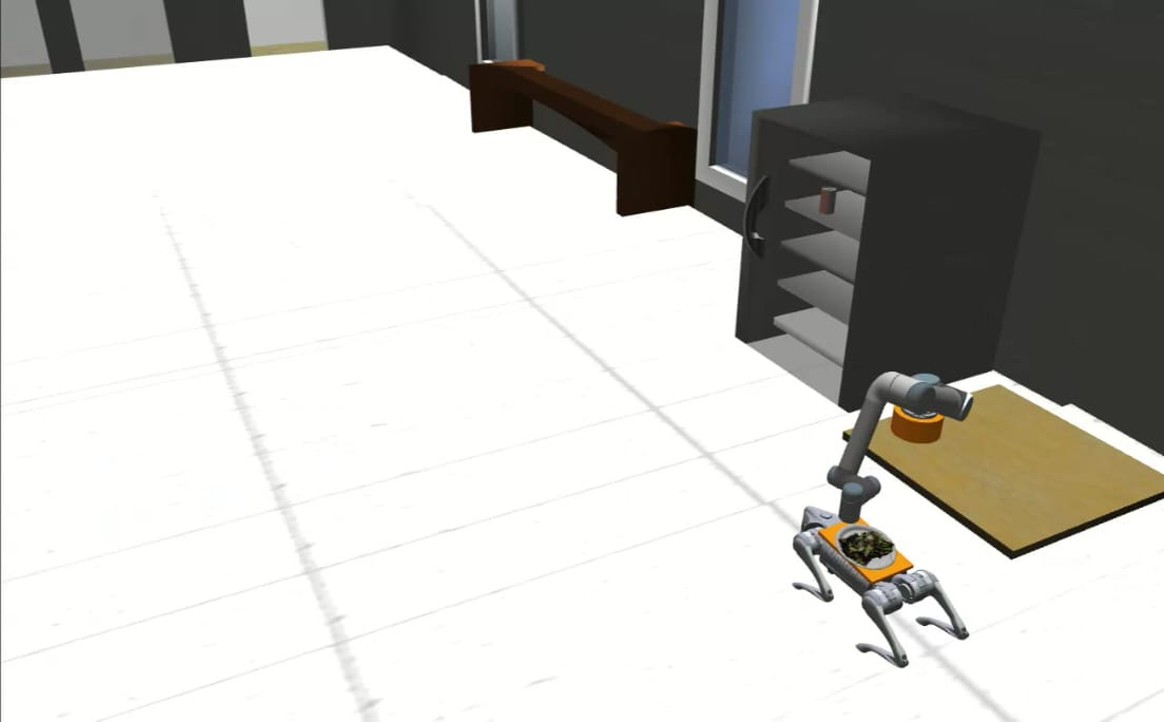}
        \caption{UR5 chef places plate on quadruped.}
        \label{fig:hospital_C}
    \end{subfigure}\hfill
    \begin{subfigure}[t]{0.48\linewidth}
        \includegraphics[width=\linewidth]{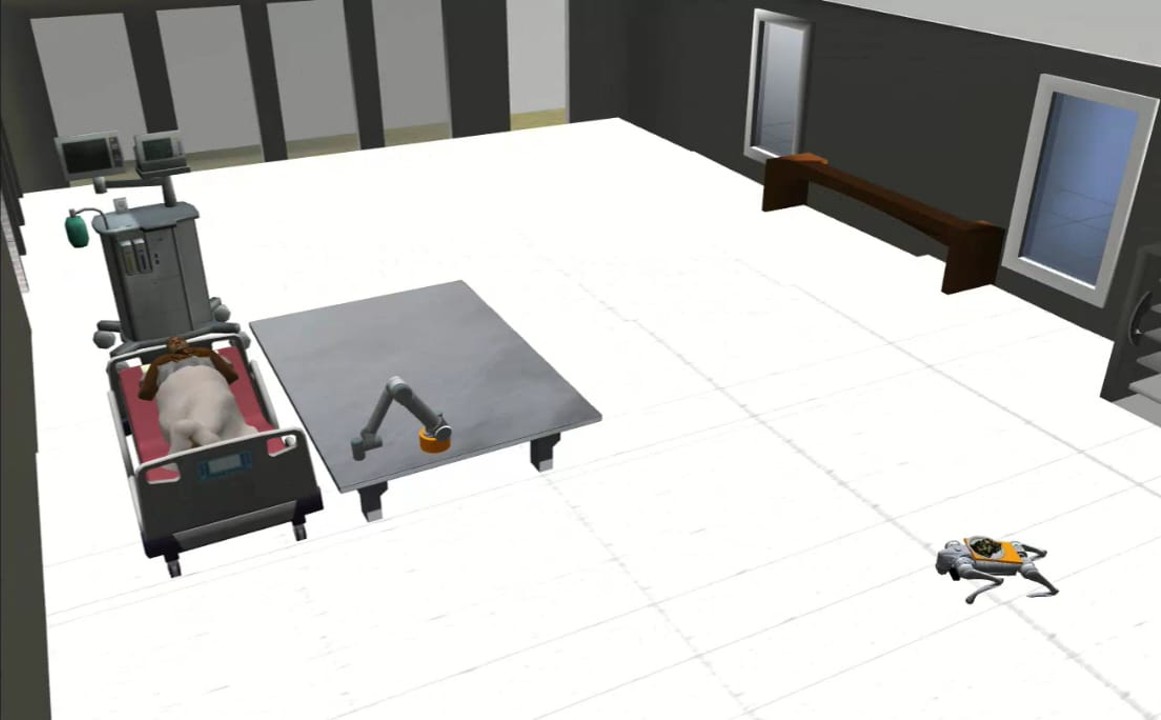}
        \caption{Quadruped moves toward patient.}
        \label{fig:hospital_D}
    \end{subfigure}

    \begin{subfigure}[t]{0.48\linewidth}
        \includegraphics[width=\linewidth]{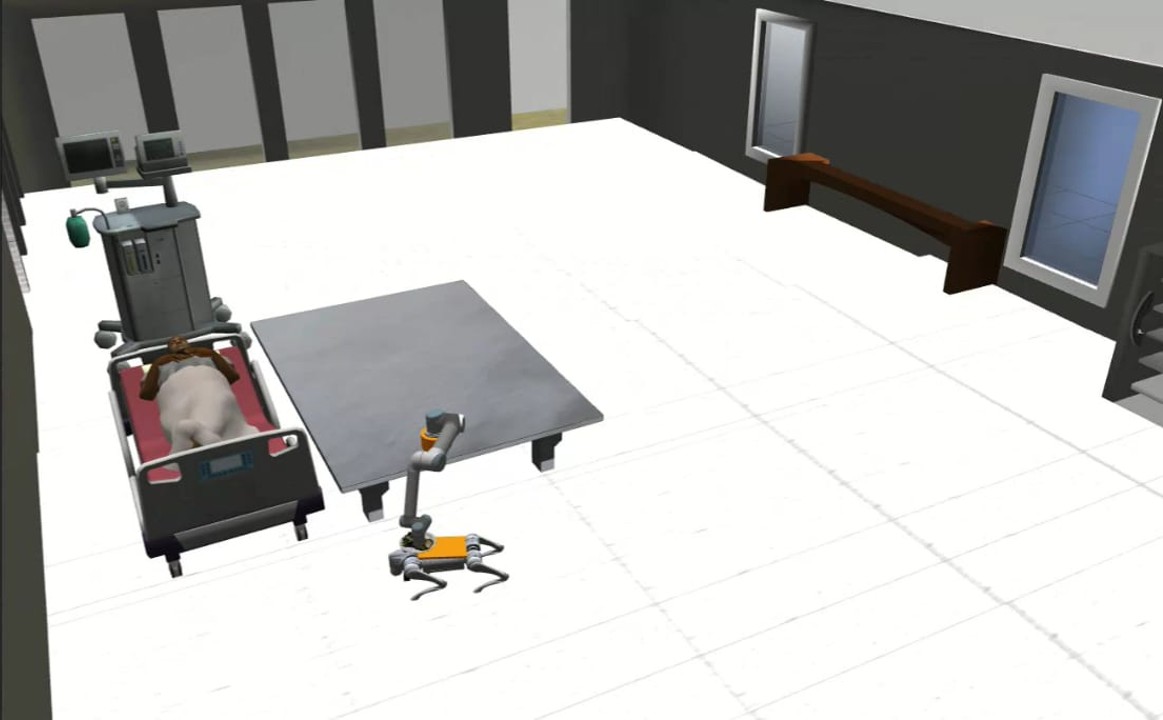}
        \caption{UR5 helper takes plate.}
        \label{fig:hospital_E}
    \end{subfigure}\hfill
    \begin{subfigure}[t]{0.48\linewidth}
        \includegraphics[width=\linewidth]{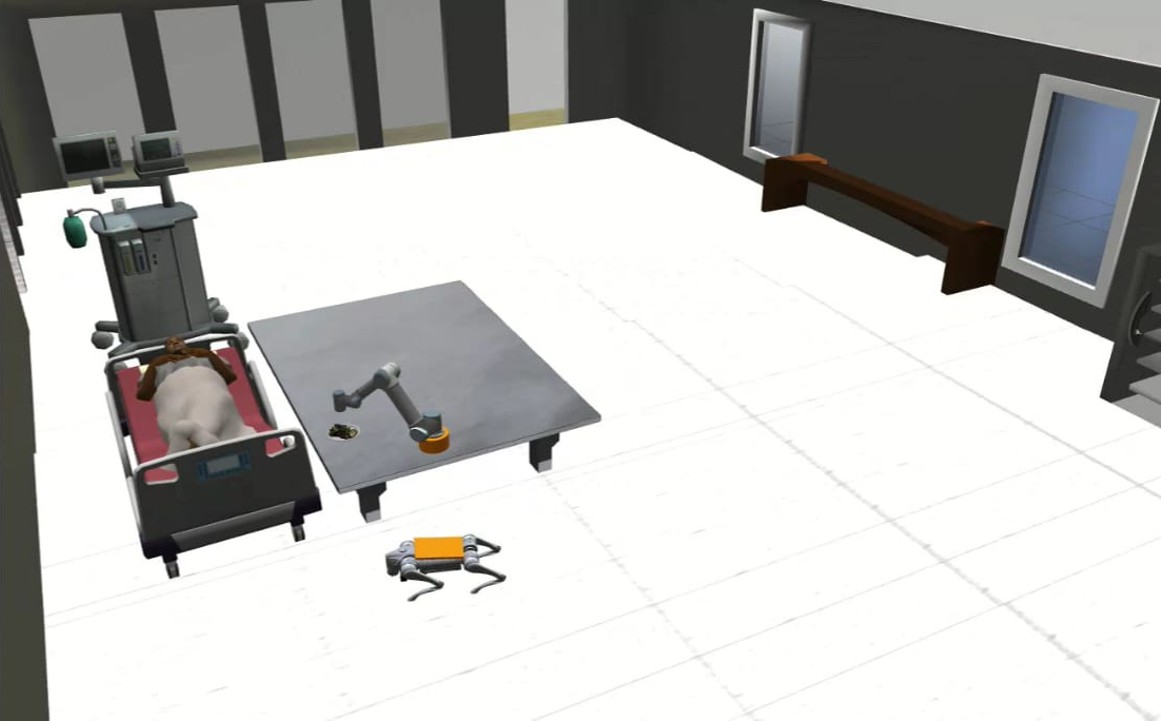}
        \caption{UR5 places plate on table.}
        \label{fig:hospital_F}
    \end{subfigure}

    \caption{Hospital scenario: interpreting the human prompt ``I am hungry'' and coordinating UR5 arms with a quadruped to deliver food.}
    \label{fig:hospital}
\end{figure}

\begin{figure}[h]
    \centering
    \setlength{\belowcaptionskip}{-22pt}
    \captionsetup[subfigure]{font=footnotesize,aboveskip=0pt,belowskip=0pt}

    \begin{subfigure}[t]{0.48\linewidth}
        \includegraphics[width=\linewidth]{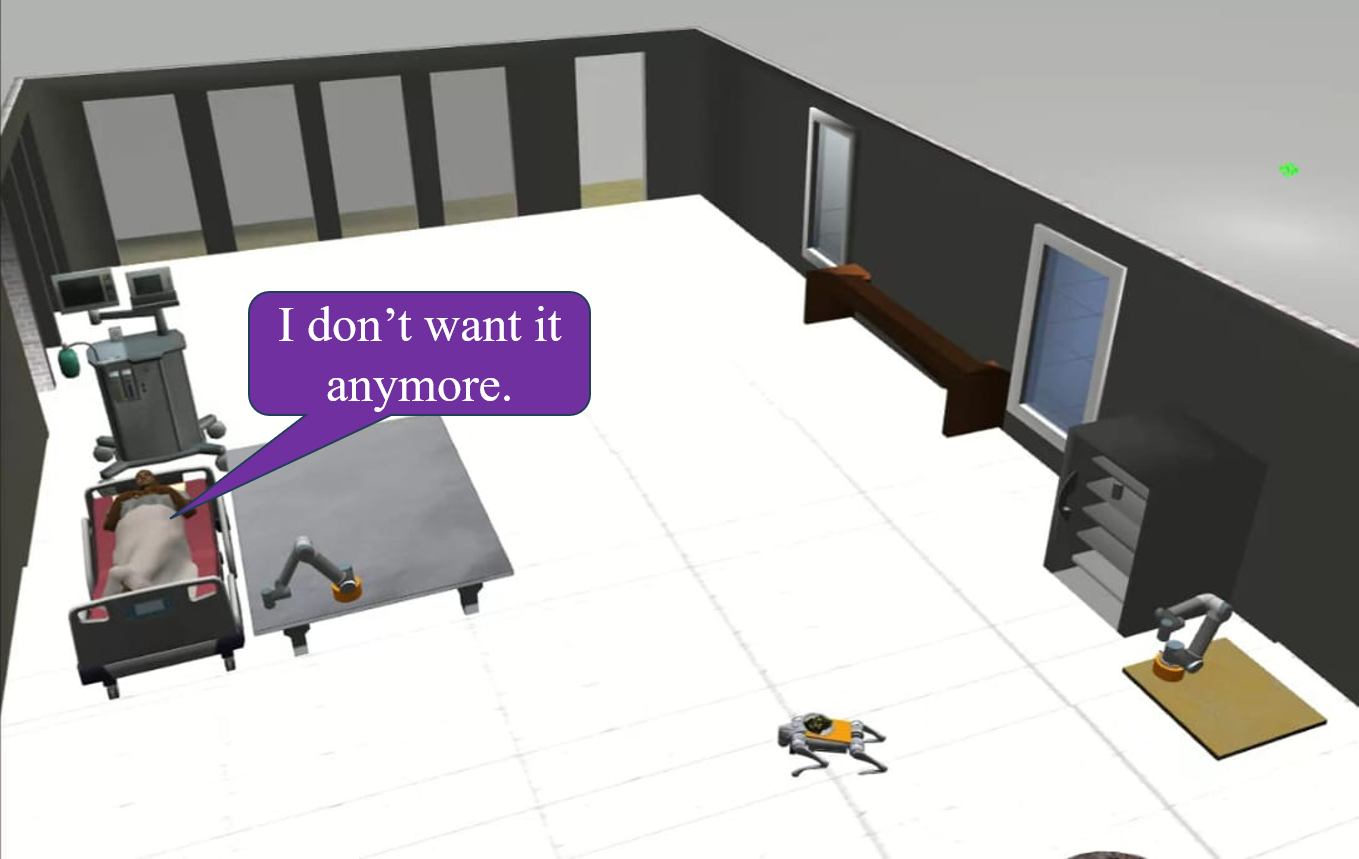}
        \caption{Patient refuses food.}
        \label{fig:hospital_event_E}
    \end{subfigure}\hfill
    \begin{subfigure}[t]{0.48\linewidth}
        \includegraphics[width=\linewidth]{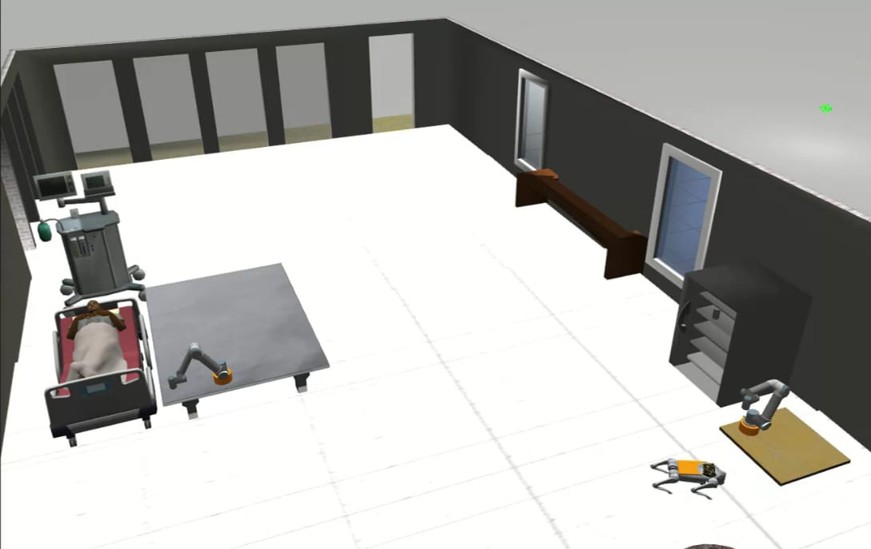}
        \caption{Quadruped returns food to table.}
        \label{fig:hospital_event_F}
    \end{subfigure}

    \caption{Hospital scenario: demonstration of anytime human interruption, with the quadruped adapting its plan based on updated intent.}
    \label{fig:hospital_event}
\end{figure}

In order to show relevance of CoMuRoS to diverse and real problems, simulations were performed in Gazebo environment for two scenarios Hospital and Disaster Relief. 
\subsubsection{Task Allocation in Disaster Relief Scenario}
\begin{figure}[h]
    \centering
    \setlength{\belowcaptionskip}{-20pt}
    \captionsetup[subfigure]{font=footnotesize,aboveskip=0pt,belowskip=0pt}

    \begin{subfigure}[t]{0.48\linewidth}
        \includegraphics[width=\linewidth]{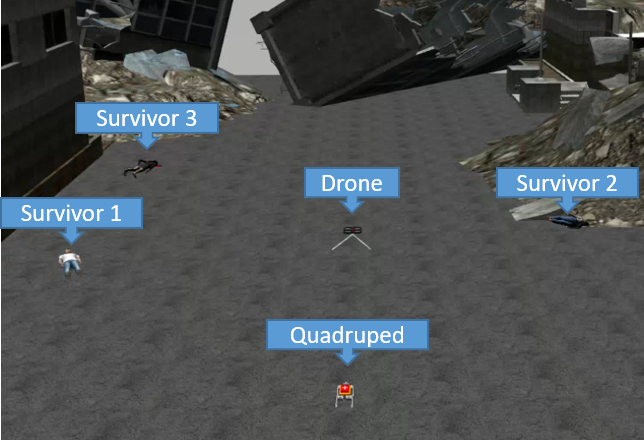}
        \caption{Overview of scene.}
        \label{fig:drone_A}
    \end{subfigure}\hfill
    \begin{subfigure}[t]{0.48\linewidth}
        \includegraphics[width=\linewidth]{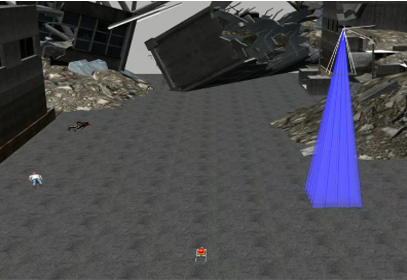}
        \caption{Drone starts surveillance.}
        \label{fig:drone_B}
    \end{subfigure}

    \begin{subfigure}[t]{0.48\linewidth}
        \includegraphics[width=\linewidth]{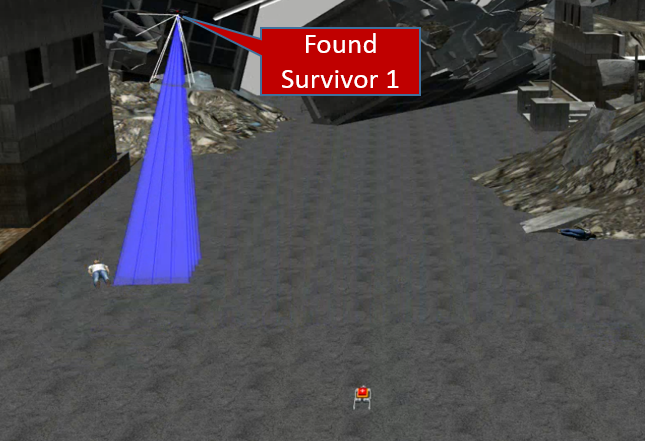}
        \caption{Drone finds a survivor and reports coordinates.}
        \label{fig:drone_C}
    \end{subfigure}\hfill
    \begin{subfigure}[t]{0.48\linewidth}
        \includegraphics[width=\linewidth]{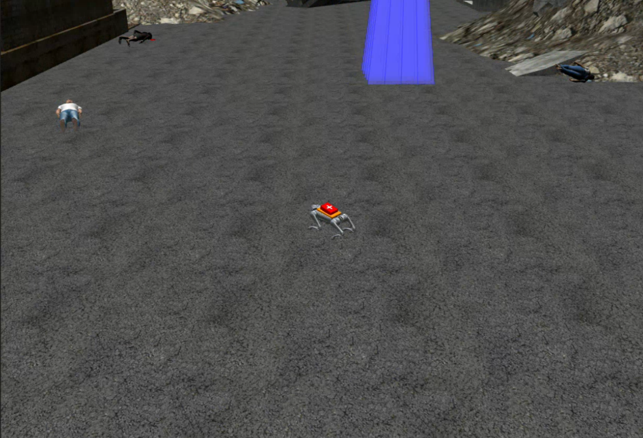}
        \caption{Quadruped moves to survivor 1 with first aid kit.}
        \label{fig:drone_D}
    \end{subfigure}

    \begin{subfigure}[t]{0.48\linewidth}
        \includegraphics[width=\linewidth]{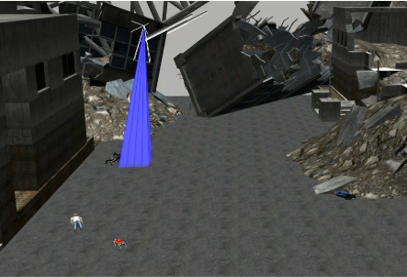}
        \caption{Drone finds survivors 2 and 3 and reports coordinates.}
        \label{fig:drone_E}
    \end{subfigure}\hfill
    \begin{subfigure}[t]{0.48\linewidth}
        \includegraphics[width=\linewidth]{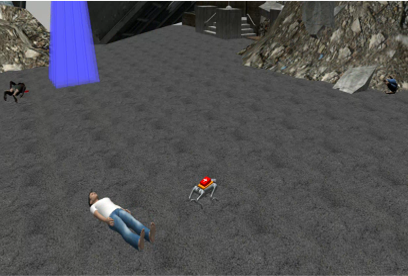}
        \caption{Quadruped reaches survivor 1 then moves to next.}
        \label{fig:drone_F}
    \end{subfigure}

    \caption{Demonstration of task allocation, classification, and replanning in a disaster relief scenario.}
    \label{fig:drone}
\end{figure}
In the disaster relief scenario (Fig.~\ref{fig:drone}), a Go2 quadruped and a drone are tasked with finding survivors and delivering first aid kits. The user command, ``Find all survivors and deliver first aid kits to them,'' is classified as an independent task: the drone searches while the quadruped delivers. When the drone detects a survivor, it raises a relevant event, triggering replanning. The Task Manager provides the survivor’s coordinates to the quadruped while the drone continues scanning.  
\vspace{-1mm}
\subsubsection{Understanding Human Intentions: Multirobot Food Delivery}
In the hospital scenario (Fig.~\ref{fig:hospital_A}), a Unitree Go2 quadruped and two UR5 arms collaborate to deliver food after the user says, ``I am hungry.'' The chef arm loads the plate, the quadruped carries it, and the helper arm serves it (Figs.~\ref{fig:hospital_B}–\ref{fig:hospital_F}). CoMuRoS also supports anytime human input: in Fig.~\ref{fig:hospital_event}, when the user interrupts with ``I don't want it anymore,'' this is classified as a relevant event, and replanning directs the quadruped to return the food. This demonstrates the system’s ability to interpret indirect human intent and adapt plans in real time.  

The videos, user interactions, code and textual dataset are available at \href{https://CoMuRoS.github.io}{CoMuRoS.github.io}.

\section{Generalization results across tasks, scenarios and LLMs}
\begin{figure}[!h]
    \centering
    \setlength{\belowcaptionskip}{-10pt}
    \includegraphics[width=1\columnwidth, trim=30 0 70 0, clip]{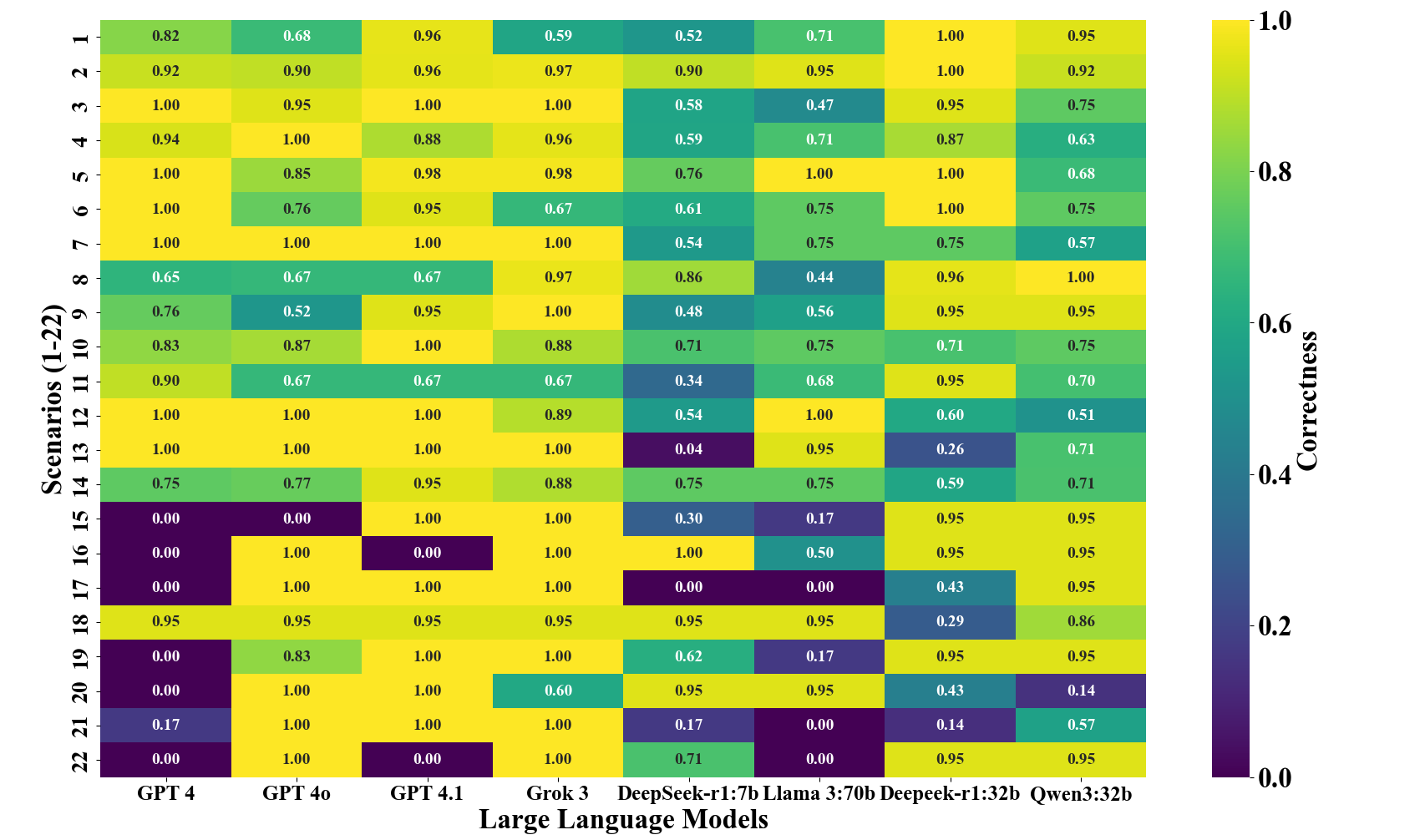}
    \caption{Correctness scores across scenarios(1-22) across different LLMs(GPT 4, GPT 4o, GPT 4.1, Grok 3, DeepSeek-r1:7b, Llama 3:70b, DeepSeek-r1:32b and Qwen3:32b).The 22 scenarios are 1. reforestation, 2. baggage handling, 3. asteroid mining, 4. beach cleaning, 5. border patrol, 6. construction, 7. crop monitoring, 8. gardening, 9. highway cleaning, 10. hospital, 11. house cleaning, 12. mine detection, 13. office transport, 14. school assistants, 15. beam transport, 16. wheel change, 17. wall lift, 18. fire rescue, 19. solar panel install, 20. lift stretcher, 21. mirror mounting, 22. boat towing.}
\label{fig:heatmap-correctness}
\end{figure}
CoMuRoS architecture's high-level planning capability was evaluated using a manually curated textual benchmark dataset for generalization across 22 scenarios, with three tasks evaluated by eight different LLMs (APIs and local). Each task is evaluated five times. 
\begin{figure}[!h]
    \centering  
    \setlength{\belowcaptionskip}{-15pt}
    \includegraphics[width=\linewidth, trim=0 0 0 0, clip]{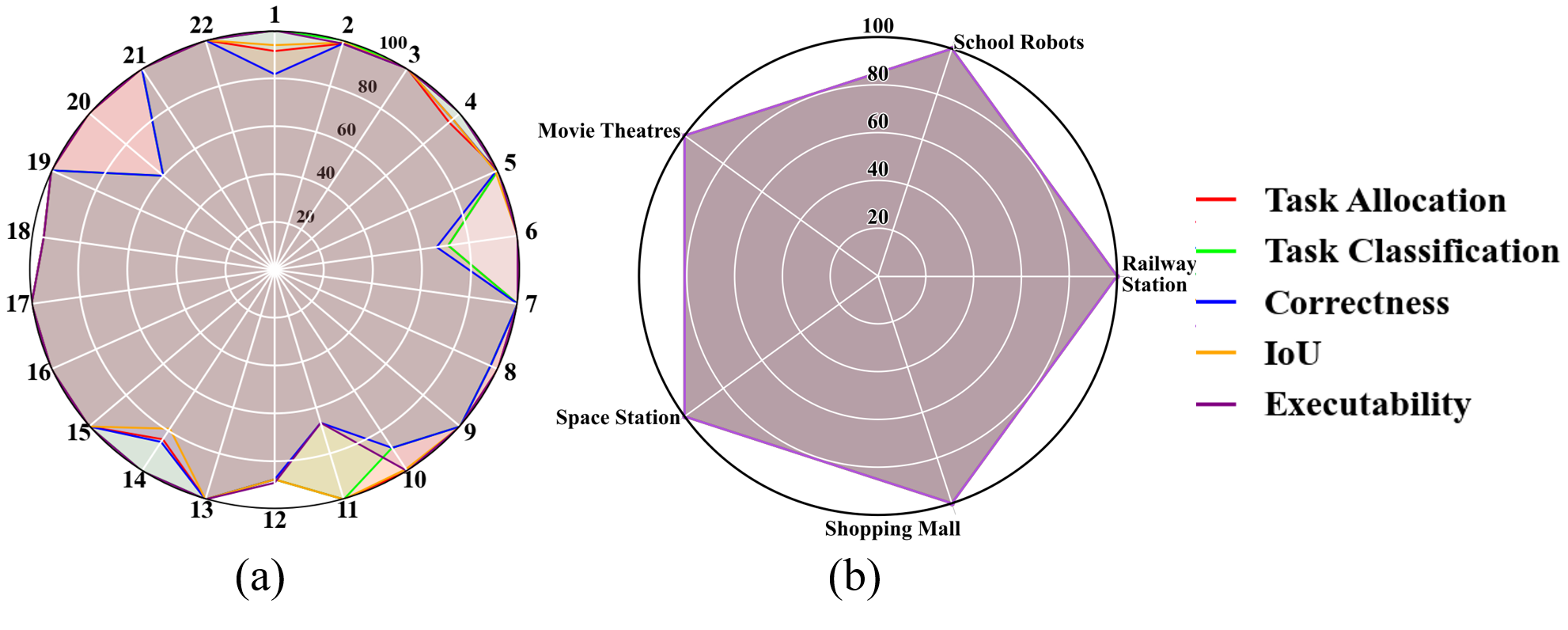}
    \caption{(a) Per-metric performance across 22 scenarios(mentioned in Fig. \ref{fig:heatmap-correctness}) for Grok 3 . Each axis is a scenario, with values showing average metric scores (0–1) across all tasks and iterations. (b) Evaluation for replanning in different tasks for Grok 3.}
\label{fig:metric-radar}
\end{figure}

\begin{figure}[!h]
    \centering
    \setlength{\belowcaptionskip}{-20pt}
    \includegraphics[width=0.8\linewidth, trim=20 0 10 0, clip]{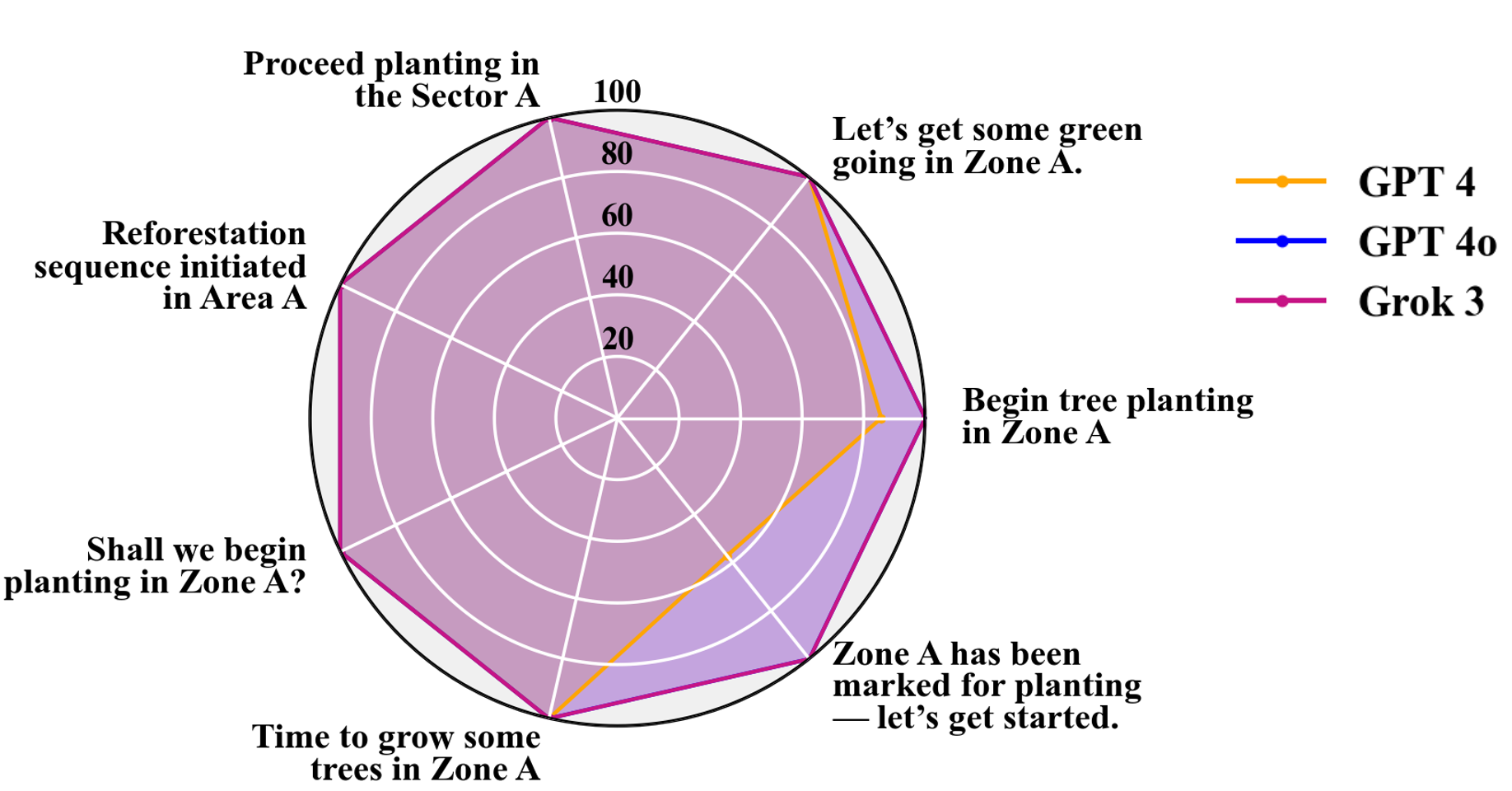}
    \caption{Correctness scores different LLMs (GPT 4, GPT 4o and Grok 3) for different phrasings of semantically identical prompts.}
    \label{fig:same_meaning_spider}
\end{figure}

As seen in Fig. \ref{fig:heatmap-correctness} CoMuRoS performs better using Grok 3 than other LLMs. CoMuRoS has an average correctness of 0.91 while using Grok 3 (averaged across 22 scenarios having 3 tasks each, and each task was evaluated 5 times).

Similarly, the average values for Grok 3 are: TA = 0.96, TC = 0.96, IoU = 0.97, and Executability = 0.98. Fig.~\ref{fig:metric-radar}a shows CoMuRoS performance on these metrics across all scenarios.
A second dataset of 5 scenarios (3 tasks each) evaluates replanning using the same metrics. For each scenario, the dataset provides an initial user command and an event description. The Task Manager generates an initial plan, then replans after the event. The new plan is evaluated on the same metrics only when the initial plan is correct and the event is relevant. Each task is tested 5 times. As shown in Fig.~\ref{fig:metric-radar}b, CoMuRoS achieves a correctness of 1.0 across all 5 iterations, 3 tasks, and 5 scenarios with Grok 3.
The robustness of CoMuRoS to different phrasings of the same user prompt for task planning is also evaluated. Fig.~\ref{fig:same_meaning_spider} shows correctness scores for semantically identical prompts using GPT 4, GPT 4o, and Grok 3.
\vspace{-1mm}
\section{Conclusion}
This work presents a hierarchical multi-robot architecture, CoMuRoS, that combines Large Language Models for planning, enabling proactive and adaptive teamwork. The task manager redistributes tasks in response to events, inspired by human teams where members can signal issues and managers adapt roles. Hardware trials on heterogeneous robots and simulations show high success rates. The architecture is both modular and generalizable: keeping the static prompt and core task manager logic unchanged, and modifying only the scenario configuration file, gave consistent success across textual datasets, simulations, and hardware experiments. Future advances in foundation models promise even more flexible, human-like collaboration.

\textbf{Limitations and Future Directions:} Replanning currently ensures task completion after failures; future directions include reasoning about unrecoverable failures and pursuing partial completion to minimize loss. Further work will explore domain-specific adaptations, prompt pruning to reduce overhead, utilizing a map of the environment and advanced context management for token efficiency.

\section*{ACKNOWLEDGEMENT}
The authors thank the Government of India for supporting the first author’s research through the Prime Minister’s Research Fellowship (PMRF) Scheme.
The authors thank Mr. Aarav Shah, an undergraduate student at IIT Gandhinagar for his help with hardware setup and Ms. Saumya Karan, a PhD Student at IIT Gandhinagar for her help in dataset evaluation. Finally the authors thank IIT Gandhinagar and its donors for their unwavering support.









\bibliographystyle{IEEEtran}
\bibliography{references}

@unpublished{comuros,
  title     = {CoMuRoS: Adaptive Hierarchical Planning for Multi-Robot Teams in Dynamic Environments Using LLMs},
  author    = {Borate, Suraj and Rai B, Bhavish and Pardeshi, Vipul and Vadali, Madhu},
  note      = {Manuscript submitted to an international robotics conference},
  year      = {2025}
}

@article{brohan2022rt,
  title={Rt-1: Robotics transformer for real-world control at scale},
  author={Brohan, Anthony and Brown, Noah and Carbajal, Justice and Chebotar, Yevgen and Dabis, Joseph and Finn, Chelsea and Gopalakrishnan, Keerthana and Hausman, Karol and Herzog, Alex and Hsu, Jasmine and others},
  journal={arXiv preprint arXiv:2212.06817},
  year={2022}
}

@inproceedings{zitkovich2023rt,
  title={Rt-2: Vision-language-action models transfer web knowledge to robotic control},
  author={Zitkovich, Brianna and Yu, Tianhe and Xu, Sichun and Xu, Peng and Xiao, Ted and Xia, Fei and Wu, Jialin and Wohlhart, Paul and Welker, Stefan and Wahid, Ayzaan and others},
  booktitle={Conference on Robot Learning},
  pages={2165--2183},
  year={2023},
  organization={PMLR}
}

@article{kim2024openvla,
  title={Openvla: An open-source vision-language-action model},
  author={Kim, Moo Jin and Pertsch, Karl and Karamcheti, Siddharth and Xiao, Ted and Balakrishna, Ashwin and Nair, Suraj and Rafailov, Rafael and Foster, Ethan and Lam, Grace and Sanketi, Pannag and others},
  journal={arXiv preprint arXiv:2406.09246},
  year={2024}
}

@article{bjorck2025gr00t,
  title={Gr00t n1: An open foundation model for generalist humanoid robots},
  author={Bjorck, Johan and Casta{\~n}eda, Fernando and Cherniadev, Nikita and Da, Xingye and Ding, Runyu and Fan, Linxi and Fang, Yu and Fox, Dieter and Hu, Fengyuan and Huang, Spencer and others},
  journal={arXiv preprint arXiv:2503.14734},
  year={2025}
}

@article{barto2021reinforcement,
  title={Reinforcement learning: An introduction. by richard’s sutton},
  author={Barto, Andrew G},
  journal={SIAM Rev},
  volume={6},
  number={2},
  pages={423},
  year={2021},
  publisher={SIAM}
}

@book{spong2008robot,
  title={Robot dynamics and control},
  author={Spong, Mark W and Vidyasagar, Mathukumalli},
  year={2008},
  publisher={John Wiley \& Sons}
}

@article{aeronautiques1998pddl,
  title={Pddl—the planning domain definition language},
  author={Aeronautiques, Constructions and Howe, Adele and Knoblock, Craig and McDermott, ISI Drew and Ram, Ashwin and Veloso, Manuela and Weld, Daniel and Sri, David Wilkins and Barrett, Anthony and Christianson, Dave and others},
  journal={Technical Report, Tech. Rep.},
  year={1998}
}

@article{fikes1971strips,
  title={STRIPS: A new approach to the application of theorem proving to problem solving},
  author={Fikes, Richard E and Nilsson, Nils J},
  journal={Artificial intelligence},
  volume={2},
  number={3-4},
  pages={189--208},
  year={1971},
  publisher={Elsevier}
}

@article{wang2024dart,
  title={Dart-llm: Dependency-aware multi-robot task decomposition and execution using large language models},
  author={Wang, Yongdong and Xiao, Runze and Kasahara, Jun Younes Louhi and Yajima, Ryosuke and Nagatani, Keiji and Yamashita, Atsushi and Asama, Hajime},
  journal={arXiv preprint arXiv:2411.09022},
  year={2024}
}

@inproceedings{kannan2024smart,
  title={Smart-llm: Smart multi-agent robot task planning using large language models},
  author={Kannan, Shyam Sundar and Venkatesh, Vishnunandan LN and Min, Byung-Cheol},
  booktitle={2024 IEEE/RSJ International Conference on Intelligent Robots and Systems (IROS)},
  pages={12140--12147},
  year={2024},
  organization={IEEE}
}

@article{liu2024coherent,
  title={Coherent: Collaboration of heterogeneous multi-robot system with large language models},
  author={Liu, Kehui and Tang, Zixin and Wang, Dong and Wang, Zhigang and Li, Xuelong and Zhao, Bin},
  journal={arXiv preprint arXiv:2409.15146},
  year={2024}
}

@inproceedings{shah2023lm,
  title={Lm-nav: Robotic navigation with large pre-trained models of language, vision, and action},
  author={Shah, Dhruv and Osi{\'n}ski, B{\l}a{\.z}ej and Levine, Sergey and others},
  booktitle={Conference on robot learning},
  pages={492--504},
  year={2023},
  organization={PMLR}
}

@inproceedings{huang2022language,
  title={Language models as zero-shot planners: Extracting actionable knowledge for embodied agents},
  author={Huang, Wenlong and Abbeel, Pieter and Pathak, Deepak and Mordatch, Igor},
  booktitle={International conference on machine learning},
  pages={9118--9147},
  year={2022},
  organization={PMLR}
}

@inproceedings{mandi2024roco,
  title={Roco: Dialectic multi-robot collaboration with large language models},
  author={Mandi, Zhao and Jain, Shreeya and Song, Shuran},
  booktitle={2024 IEEE International Conference on Robotics and Automation (ICRA)},
  pages={286--299},
  year={2024},
  organization={IEEE}
}

@inproceedings{chen2024scalable,
  title={Scalable multi-robot collaboration with large language models: Centralized or decentralized systems?},
  author={Chen, Yongchao and Arkin, Jacob and Zhang, Yang and Roy, Nicholas and Fan, Chuchu},
  booktitle={2024 IEEE International Conference on Robotics and Automation (ICRA)},
  pages={4311--4317},
  year={2024},
  organization={IEEE}
}

@article{firoozi2025foundation,
  title={Foundation models in robotics: Applications, challenges, and the future},
  author={Firoozi, Roya and Tucker, Johnathan and Tian, Stephen and Majumdar, Anirudha and Sun, Jiankai and Liu, Weiyu and Zhu, Yuke and Song, Shuran and Kapoor, Ashish and Hausman, Karol and others},
  journal={The International Journal of Robotics Research},
  volume={44},
  number={5},
  pages={701--739},
  year={2025},
  publisher={SAGE Publications Sage UK: London, England}
}

@article{zhang2023building,
  title={Building cooperative embodied agents modularly with large language models},
  author={Zhang, Hongxin and Du, Weihua and Shan, Jiaming and Zhou, Qinhong and Du, Yilun and Tenenbaum, Joshua B and Shu, Tianmin and Gan, Chuang},
  journal={arXiv preprint arXiv:2307.02485},
  year={2023}
}

@inproceedings{karapetyan2017efficient,
  title={Efficient multi-robot coverage of a known environment},
  author={Karapetyan, Nare and Benson, Kelly and McKinney, Chris and Taslakian, Perouz and Rekleitis, Ioannis},
  booktitle={2017 IEEE/RSJ International Conference on Intelligent Robots and Systems (IROS)},
  pages={1846--1852},
  year={2017},
  organization={IEEE}
}

@article{alonso2017multi,
  title={Multi-robot formation control and object transport in dynamic environments via constrained optimization},
  author={Alonso-Mora, Javier and Baker, Stuart and Rus, Daniela},
  journal={The International Journal of Robotics Research},
  volume={36},
  number={9},
  pages={1000--1021},
  year={2017},
  publisher={SAGE Publications Sage UK: London, England}
}

\end{document}